# AI-Based Measurement of Innovation: Mapping Expert Insight into Large Language Model Applications

Robin Nowak[*], Patrick Figge, Carolin Häussler

*University of Passau*

## Abstract

Measuring innovation often relies on context-specific proxies and on expert evaluation. Hence, empirical innovation research is often limited to settings where such data is available. We investigate how large language models (LLMs) can be leveraged to overcome the constraints of manual expert evaluations and assist researchers in measuring innovation. We design an LLM framework that reliably approximates domain experts' assessment of innovation from unstructured text data. We demonstrate the performance and broad applicability of this framework through two studies in different contexts: (1) the innovativeness of software application updates and (2) the originality of user-generated feedback and improvement ideas in product reviews. We compared the performance (F1-score) and reliability (consistency rate) of our LLM framework against alternative measures used in prior innovation studies, and to state-of-the-art machine learning- and deep learning-based models. The LLM framework achieved higher F1-scores than the other approaches, and its results are highly consistent (i.e., results do not change across runs). This article equips R&D personnel in firms, as well as researchers, reviewers, and editors, with the knowledge and tools to effectively use LLMs for measuring innovation and evaluating the performance of LLM-based innovation measures. In doing so, we discuss, the impact of important design decisions—including model selection, prompt engineering, training data size, training data distribution, and parameter settings—on performance and reliability. Given the challenges inherent in using human expert evaluation and existing text-based measures, our framework has important implications for harnessing LLMs as reliable, increasingly accessible, and broadly applicable research tools for measuring innovation.

---

[*] Please send correspondence to robin.nowak@uni-passau.de.

The project was funded by the Deutsche Forschungsgemeinschaft (DFG, German Research Foundation) - project number 443732464.

## 1. Introduction

Innovation scholars have long been interested in measuring to what extend an activity or outcome is 'innovative' (Griliches 1990, Smith 2005, Verhoeven et al. 2016, Gault 2018, Bellstam et al. 2021). This is inherently difficult as innovative activity reflects a broad spectrum of activities which range from developing new products, business models, and processes to adopting novel organizational structures or reconfiguring existing systems to create value in new ways (Schumpeter 1934). Capturing this multidimensional nature requires varied metrics and approaches, making standardized measurement challenging. Furthermore, discerning which activity truly qualifies as 'an innovation' typically requires the specialized knowledge of domain experts (Rosenkopf and McGrath 2011). As a result, quantitative studies focus on contexts where expert-assessed data is publicly available. For instance, patent-related proxies are widely used because the patent application process involves a rigorous examination of an invention's novelty by patent examiners at patent offices (Griliches 1990). Similarly, in pharmaceutical contexts, clinical trials and new drug approvals offer a rich setting to capture expert-evaluated output from innovative activity (e.g., Haeussler and Assmus 2021, Krieger 2021). However, in many other domains, accessing such data or expert evaluators is challenging, limiting possibilities to capture innovative activity and risking a biased perspective that favors contexts where expert-based measures are readily available (e.g., Moser 2012). Alternatively, innovation researchers have to rely on the support of experts recruited to assist in a specific project to manually assess and code innovation activities and outcomes, which severely limits sample size.

In response, scholars have begun to adopt natural language processing (NLP) and machine learning (ML) techniques to create alternatives for constructing quantitative variables based on textual data (Gentzkow et al. 2019, Miric, Jia, et al. 2023). While often and successfully implemented (Wang et al. 2018, e.g., Bellstam et al. 2021, Guzman and Li 2023, Miric, Ozalp, et al. 2023), these techniques come with substantial limitations. First, many NLP techniques depend on unambiguous, structured language for accurate performance (e.g., Loughran and McDonald 2011). Second, while 'classical' ML approaches can tolerate linguistic noise, they generally rely on large sets of expert-labeled training data to yield reliable outcomes (Hirschberg and Manning 2015). In contexts where the nature of the data and the construct to be measured involves understanding nuanced, multi-dimensional information, these approaches struggle to perform on par with human expert evaluation (Gentzkow et al. 2019). Third, applying these approaches typically demands proficiency in programming and considerable methodological expertise in, for instance, ML techniques.

Large language models (LLMs) may enable us to overcome some of these limitations. Built upon the transformer architecture (Vaswani et al. 2017) and pre-trained on vast text corpora, LLMs are able to accurately capture many of the subtleties of human language. The latest generation of LLMs bring enhanced predictive capabilities while requiring little to no additional data to perform tasks (Brown et al. 2020). They can be instructed by written or spoken human language, so-called prompts, that determine models' output generation. Outputs are often remarkably human-like across various scenarios (Jones and Bergen 2024), with LLMs even capable of adopting different viewpoints in experimental settings (Boussioux et al. 2024, Doshi et al. 2024). These advancements might allow the analysis of text data in novel ways: LLMs may also be able to capture and mimic the experts' evaluation process, holding the potential to scale expert evaluation with unprecedent performance. This way, they can democratize the analysis of unstructured text data to a wider audience of scholars by lowering barriers to use, e.g., by having little data requirements, being easy to operate, and demanding little programming or ML expertise.

In this paper, we introduce a framework for how to develop an LLM application to measure innovation-related constructs based on unstructured text data that mimics how an expert would have evaluated these constructs. Recognizing that evaluations yield the most accurate results when informed by specialized expertise (Csaszar and Eggers 2013, Böttcher and Klingebiel 2025), our LLM framework involves a structured process to capture the evaluation resulting from experts' reasoning and decision-making steps. First, we inductively identify content dimensions that indicate innovativeness in our contexts, which we then validate with domain experts to establish the foundation for prompt design. Second, we move beyond the default capabilities of the pre-trained model by fine-tuning it with expert-labeled training data. Third, we address the reliability of results in repeated analyses by discussing variations in design decisions including prompting, model selection, fine-tuning data, and parameter settings.

We demonstrate the performance and broad applicability of this framework through two studies. In study 1, we measure the innovativeness of outcomes from product development activities in the context of software application updates. In study 2, we assess the originality of user suggestions for product development activities in the context of product reviews. Both demonstration studies traditionally require methods relying heavily on human-coded labels due to the inherent complexity of the constructs being analyzed. Moreover, in both studies, the textual data serving as the basis for evaluation is highly nuanced, context-specific, and lacks clear structures or standardized guidelines, making outcomes such as updates or reviews subject to subtle linguistic variations.

In both studies, our LLM application outperformed (in terms of F1-scores) innovation measures recently used in scholarly work as well as established NLP techniques. When fine-tuning with expert-labeled training data is possible, LLMs achieve their highest performance, as this data allows for targeted mitigation of biases—such as those stemming from structural misclassifications or failed classifications of texts—and ensures high reliability independent of design decisions. But even without expert-labeled training data, LLMs still achieve strong performance and consistency. Overall, our findings reveal that our LLM framework is highly adjustable to both the measurement objective and context, provided that specialized human experts are available to clearly define measurement criteria and assess its accuracy on a training dataset.

In further analyses, we systematically explore and vary a series of design decisions to evaluate their impact, aiming to derive recommendations on how to use LLMs as research tools. First, while more recent LLMs generally deliver improved performance, GPT models emerge as best-in-class, and fine-tuning equalizes performance differences among models. Second, when prompts are carefully crafted and validated by experts to guide through data structures and classification patterns, additional prompt engineering offers only marginal performance gains. For instance, few-shot prompting provides slight performance improvements for default LLMs but has little impact on fine-tuned LLMs and may even risk biasing measurement outcomes by overly weighting certain data points that correspond to the selected examples. Third, increasing the temperature reduces reliability with no performance gains. Finally, while training data size is essential for many NLP approaches' performance, fine-tuned LLMs, particularly GPT models, perform robustly even with small training data sets. Moreover, strategically managing class distribution through methods such as oversampling can enhance the identification of underrepresented content dimensions.

Our work makes three key contributions. First, we add to an emerging line of research that applies LLMs as research tools to construct variables based on text data (Fyffe et al. 2023, Just et al. 2023, Cheng et al. 2024, Masclans-Armengol et al. 2024, Ganguli et al. 2024, Carlson and Burbano 2025, de Kok 2025). Our framework presents a low-barrier approach for researchers to access and apply reliable expert-level evaluations in various innovation domains. Specifically, we provide a systematic process that guides researchers and R&D personnel in firms through a structured prompt design, fine-tuning, and validation of a pre-trained LLM informed by validated domain-specific content dimensions to create measures of interest. Researchers wishing to adapt our framework to their own context need only to define the construct of interest by identifying key indicators, collaborate with few specialized experts for validation, and then follow our detailed guidelines on prompt design, model selection, and evaluation.

Second, our discussion on the implications of design decisions informs researchers on how variations in prompt engineering, fine-tuning strategies, and parameter settings can affect performance and reliability (Carlson and Burbano 2025). This information enables researchers and R&D personnel in firms to tailor their LLM applications to balance optimal outcomes with practical constraints, such as time spent and costs of fine-tuning, and supports the scholarly community in interpreting, evaluating, and extending LLM-based work (Dell'Acqua et al. 2023, Otis et al. 2024, Boussioux et al. 2024). Moreover, this article contributes to equipping reviewers and editors with the knowledge on how evaluate the use of LLMs in measuring innovation (de Kok 2025).

Third, our demonstration studies introduce novel, high-quality measures that enable more precise estimations of the type and degree of innovation in software application development and product review contexts. They offer researchers and R&D personnel in firms a robust alternative to traditional proxies and human expert assessments. The respective code, local models, training and validation datasets, and fine-tuning approaches are all made public in an online repository[1] to support replication and adaptation for future research.

The remainder of the paper is structured as follows: In Section 2, we review the relevant literature on expert evaluation in innovation research and the application of NLP techniques for creating text-based variables. In Section 3, we describe our novel methodological framework to mirror expert insight into LLM applications. In Section 4, we present two demonstration studies that showcase the application of our framework in measuring innovation-related constructs within the contexts of software application development and product reviews. In Section 5, we discuss the critical design decisions—including prompt engineering, model selection, fine-tuning data, and temperature—and their implications for performance and reliability. Finally, Section 6 discusses the broader implications of our findings and outlines avenues for future research.

## 2. Background

### 2.1. Expert Evaluation in Innovation Research

The increasing availability of text data has unlocked new opportunities to gain insights into idea development and innovative activity. On the one hand, organizations can now easily access broad pools of ideas and suggestions from several sources, which allow them to source novel solutions to innovation challenges from both internal and external contributors (Jeppesen and Lakhani 2010, Boudreau et al. 2011). On the other hand, such organizations themselves generate much text data internally, providing opportunities to detect patterns in their innovation activities beyond what is captured by traditional measures like R&D intensity or patent-related outputs. In particular, this data provides insights into broader innovation processes while also capturing contexts that do not rely on R&D expenditures and patenting (Bellstam et al. 2021).

---

[1] All data, code, and supplementary materials needed to reproduce our results are openly accessible in the project's GitHub repository (https://github.com/robi979/AI-Innovation-Measurement). We also provide a lightweight 'application package' containing the essential code and example notebooks that show how to run our framework directly through common LLM-provider APIs, enabling rapid integration without downloading the full replication bundle.

Important traditional measures are by legal definition related to inventiveness. Patents, in particular, undergo a rigorous examination process to ensure novelty, non-obviousness and industrial applicability (Title 35 United States Code), relying on specialized, highly-skilled professionals such as patent examiners of patent offices to evaluate technical details and legal requirements. Extending this level of rigor to other spheres of innovation is challenging. Unlike patents, many forms of innovation lack formalized frameworks, legal structures, or dedicated expert evaluators. Still, reliable construct measurement often demands a clear understanding of the construct, deep domain knowledge and well-developed cognitive frameworks to comprehend, organize, and interpret construct-relevant information—expertise that is both scarce and difficult to scale.

Even when experts are available, scaling their evaluation is inherently constrained by time or budget constraints and human cognitive limits. Manually processing large volumes of complex information and identifying meaningful patterns is tedious and time-intensive. Evaluators often struggle to manage the complexity of vast (multi-dimensional) datasets, particularly when faced with interacting or uncertain aspects that cannot be easily connected to existing knowledge structures. Psychology research highlights that when overwhelmed with too much information at once, humans fail to organize and interpret the data thematically, leading to suboptimal decision-making (Tversky and Kahneman 1974). These cognitive limits hinder the ability to process complex information effectively and impede innovation evaluation at scale.

The difficulty of processing large datasets not only constrains evaluators' ability to assess information but also narrows their scope of attention (Samuelson and Zeckhauser 1988). Faced with complex or uncertain information at scale, evaluators often resort to heuristics, reinforcing existing biases or adhering to prevailing concepts (Tversky and Kahneman 1974). Their attention space then tends to align with established knowledge structures, causing them to overlook unconventional information that fall outside their expertise (Piezunka and Dahlander 2015, Boudreau et al. 2016). For instance, evaluators may judge the value of new unfamiliar information based on prior decisions rather than its intrinsic merits, potentially explaining why established technologies persist even when superior alternatives exist (Abrahamson 1991). Taken to the extreme, this tendency to favor the status quo may result in the outright rejection of potential innovations when increased workload may overwhelm evaluators' cognitive resources.

However, thorough evaluation is essential for measuring valid constructs, particularly when leveraging the growing availability of large datasets to explore hitherto less researched contexts. Advances in NLP techniques, driven by progress in AI research, now present an opportunity to overcome constraints imposed by human cognitive limits in processing information (Korinek 2023). And in fact, first studies have demonstrated how these novel techniques can complement or even substitute human capabilities in generating measures (e.g., Choudhury et al. 2019, Miric, Jia, et al. 2023, Fyffe et al. 2023, de Kok 2025), decision-making (Doshi et al. 2024), and idea selection (Just et al. 2023), paving the way for a deeper understanding of how these tools can be effectively utilized to enhance both research and practice.

### 2.2. Towards Large Language Models for Expert Evaluation

Traditionally, scholars have operationalized variables at scale by using keyword-based methods to approximate the conceptual judgments that domain experts otherwise would have to make (e.g., Foerderer et al. 2018, Kircher and Foerderer 2023). This method involves the use of predefined keyword dictionaries that map specific terms to theoretically grounded constructs.

The approach is particularly effective when theoretical concepts and the terminology used to communicate them are well-defined and broadly recognized. However, keyword-based methods come with limitations. First, constructing comprehensive and reliable dictionaries can be challenging, especially when no related dictionary exists. The process often relies on subjective judgments, and validating dictionaries is non-trivial. Second, keyword-based approaches struggle to capture the context in which a word appears. For instance, they may fail to identify nuances such as negations or different semantic meanings depending on the surrounding text. While these approaches are computationally efficient and easily interpretable, their ability to comprehensively and reliably capture complex constructs is very limited (see Miric, Jia, et al. 2023 for a detailed discussion).

To address these issues, scholar have increasingly turned to machine learning (ML) models. ML encompasses statistical algorithms that refine their performance through iterative "learning" from data. Depending on the research objective, scholars can choose between supervised and unsupervised techniques. Supervised algorithms classify observations into predefined categories, making them well-suited for tasks where evaluators assess content according to specific patterns or develop categorical variables (e.g., Miric, Jia, et al. 2023). In contrast, unsupervised methods discover emergent themes in text data without relying on prior frameworks, making them particularly useful for exploratory or similarity-based analyses (e.g., Hannigan et al. 2019).

Effective application of ML to textual data requires transforming language into numerical representations that algorithms can process. Early approaches, such as bag-of-words models, convert text into numeric representations by counting term frequencies disregarding order and contextual information (Manning et al. 2008). For instance, popular models like TF-IDF (i.e., Term Frequency-Inverse Document Frequency) assign weights to words based on their frequency across documents, or use co-occurrence patterns to group words into latent themes as in latent topic distributions. However, these techniques often fail to capture deeper semantic relationships and struggle with comparing large numbers of documents due to their sparse, context-free representations (Turney and Pantel 2010).

Embedding-based approaches have improved on these limitations by creating numerical representations (embeddings) of words that encode their relationships in a given text corpus. Based on the idea that words appearing in similar contexts tend to have related meanings (Harris 1954), approaches (continuous models) like Word2Vec (Mikolov et al. 2013) and GloVe (Pennington et al. 2014) build on large text corpora and neural architectures to infer an embedding for every vocabulary item, analyzing how frequently they co-occur in a text corpus. Each word can then be represented as a vector based on all surrounding words (or vice versa). This results in vectors where semantically similar words are geometrically close, while unrelated words are farther apart. However, the embeddings generated by continuous models are static, meaning each word is assigned the same numerical representation regardless of the context in which it appears in downstream tasks. Simply put, the word "launch" would always have the same vector based on the initial text corpus, regardless of whether it now refers to "releasing a new product" or "propelling a rocket into space".

Dynamically adjusting a word's representation based on its meaning and context has become possible with the revolutionary transformer architectures (transformers). Transformers are deep neural networks that include positional encodings of each word's relative position in the text document and a self-attention mechanism to produce rich contextual embeddings (Vaswani et al. 2017). In essence, self-attention allows the model to weigh surrounding words differently when updating a word's representation, so the same word can have varying vector representations depending on its context. This breakthrough has enabled universal text representations across massive corpora, opening unprecedented avenues for capturing and analyzing the semantic content in text data.

Transformer-based models (transformer models) have excelled in a wide range of NLP tasks that previously required human expertise to achieve sufficient performance. Early pre-trained language models (PLM) such as BERT (Devlin et al. 2019) and its derivatives rely on an encoder-based transformer architecture. The encoder processes the input text holistically, capturing intricate relationships between words by producing dynamic, context-aware representations. This process results in high-quality embeddings that form the foundation for tasks such as text classification, sentiment analysis, named entity recognition, and text similarity—often achieving or even surpassing human expert-level performance.

More recently, LLMs, which are typically built on decoder-based transformer architectures, have attracted much attention. Models like GPT-3 (Brown et al. 2020) and its successors generate text by predicting the next (sub)word (token) from its preceding (sub)words (tokens). This generative capability makes LLMs highly effective for tasks such as text generation, summarization, and conversational artificial intelligence, while also performing well in other NLP tasks. Unlike encoder-based models, which typically require further task-specific fine-tuning with labeled datasets, decoder-based models can be instructed via natural language prompts. In this context, a prompt serves as a conditioning mechanism that sets the stage for the model's response by providing contextual and directive information. Technically, the prompt is concatenated with the input sequence, and through the self-attention mechanism, the model weights different parts of the prompt and context accordingly, thereby guiding the generation process (Brown et al. 2020, Liu et al. 2023). 'Reacting' to prompts, they can perform tasks directly, often without additional task-specific fine-tuning, significantly reducing the barriers to their use.

The scaling of model size and pre-training data has driven significant performance improvements (Brown et al. 2020), enabling these models to exhibit emergent abilities such as answering unseen questions, performing arithmetic, and reasoning over multiple steps. This evolution not only makes fine-tuning an option for many use cases (de Kok 2025) but also provides researchers with the ability to incorporate expert knowledge into models for various applications with minimal prerequisites.

## 3. Mapping Expert Insight into Large Language Model Applications

Evaluations in innovation contexts often place a strong emphasis on qualitative judgements such as assessing an idea's novelty (Boudreau et al. 2016, Lane et al. 2024). Such judgments are inherently interpretive, drawing upon individual perceptions (Lane et al. 2022, Kim and DellaPosta 2022), intuition (Huang and Pearce 2015), and expertise (Boudreau et al. 2016, Li 2017), and typically lack universally accepted benchmarks at the time of evaluation. Hence, selecting appropriate evaluators is essential for high evaluation quality and reliability.

Research on innovation evaluation indicates that uncertainty in evaluations is minimized when domain experts can be involved (Csaszar and Eggers 2013, Boudreau et al. 2016, Böttcher and Klingebiel 2025). Through years of learning and deliberate practice, experts develop distinctive know-how that enables them to, often intuitively, recognize intricate patterns and extract meaningful insights from complex information (Kahneman et al. 1982, Johnson et al. 1982, Ericsson and Smith 1991). As a result, they exhibit better judgement, allowing them to assess tasks and contextual cues more accurately than nonexperts and ultimately make more informed evaluations in their domain.

In this study, we aim to integrate the distinctive know-how of domain experts into LLM-based applications, so that researchers can apply them as reliable tools for approximating expert evaluation. Although LLMs excel at extracting insights from large datasets and can generate assessments that are consistent with human judgment (Brown et al. 2020), it is essential to recognize that they operate on statistical pattern recognition rather than genuine logical reasoning or expertise-based intuition (Mitchell and Krakauer 2023). This reliance on probabilistic modeling often results in variability in outputs across runs. To address this, we suggest a systematic mapping of the cues and implicit criteria that experts draw upon in their judgements and evaluations. By explicitly codifying these aspects in the input options of an LLM, it becomes possible to steer its internal probabilities, enabling outputs to more closely mirror expert-level reasoning and assessments (Liu et al. 2023).

We begin with outlining a stylized framework to discern cues indicative of expert judgement (Einhorn 1974). This involves initially *identifying relevant cues and information* by clarifying the evaluation objective and criteria and engaging with the available data to selectively isolate evaluation-relevant cues while filtering out extraneous information. Next, the identified cues are *organized into clusters or dimensions* by contextualizing each through its connection to existing knowledge. This allows to *assess the amount of each cue* as it is reflected in the data element (e.g., a certain text) being evaluated. Finally, the organized and assessed *cues are weighed* according to their importance relative to the evaluation criteria, ultimately informing the final judgement. It is important to recognize that expert judgement often appears automatic and even instantaneous due to experts' deeply internalized knowledge structures (Chase and Simon 1973, Ericsson and Smith 1991, Dane and Pratt 2007, Sukhov et al. 2021). Hence, it is unrealistic to assume that all experts consciously and explicitly follow the outlined steps. Still, this stylized framework assists in a structured identification of relevant cues indicative of expert judgement, enabling to instruct an LLM to generate outputs that better reflect expert reasoning.

To map this framework into LLM applications, we create a prompt-data-output combination that mirrors the results of expert reasoning. In other words, rather than trying to replicate the complex and often implicit nature of expert reasoning, we embed relevant evaluation cues within the LLMs' input options. LLMs operate by generating text token by token, with each token selected from a probability distribution that is conditioned on the input prompt, model tuning parameters (such as temperature settings), and, when available, fine-tuning data. When we craft prompts that mirror the described stylized LLM framework for an evaluation task—starting with a clear task definition, moving to the identification, clustering, and assessment of relevant cues, and finally weighing these cues relative to evaluation criteria—we can effectively steer the model's internal probabilities toward outputs that reflect those from expert-level reasoning (Liu et al. 2023). Furthermore, fine-tuning the model with domain-specific, labeled data can further reinforce this alignment.

-----------------------------------------

Insert Figure 1 about here

-----------------------------------------

Figure 1 illustrates the framework for designing LLM inputs to measure constructs from unstructured text data, mimicking how an expert would have processed the evaluation. The framework begins with the researcher explicitly defining the task and desired output in the prompt's opening section, aligning it with the specific measurement objectives.

Next, the LLM framework leverages inductive coding of data (Strauss and Corbin 1998, Gioia et al. 2013) until thematic and meaning saturation are reached (Hennink et al. 2017). This approach ensures that the prompt captures the critical cues that experts use to filter and prioritize information. In addition, relevance weights can be integrated directly into the prompt or applied during post-processing. This step is crucial because it aligns the model's output with expert-informed priorities and reflects the key elements of the data structure relevant for the measured construct. It is imperative that experts are directly involved or iteratively consulted during this phase to validate the final data structure, thereby laying the groundwork for a robust coding scheme—especially if fine-tuning with further training data is planned.

The final stages of the LLM framework involve validating the LLM's output against a holdout sample and, whenever possible, other established metrics. Based on performance outcomes, fine-tuning with labeled training data can be applied and design decisions may have to be adjusted. Researchers and R&D personnel need to systematically document decisions such as model selection, prompt engineering techniques (Wei et al. 2022), temperature settings (Peeperkorn et al. 2024), and fine-tuning configurations (Gao et al. 2020) to ensure reliability. For instance, tuning parameters like temperature allow control over the randomness of token selection: lower temperatures favor high-probability tokens for more deterministic outputs, while higher temperatures facilitate more varied outputs.

Understanding the implications of such specific decisions is crucial for optimizing both performance and consistency, ensuring that the model's outputs remain reliable across repeated runs. Similar to human expert evaluation, reliability measures such as intra-rater reliability (consistency within individual repeated assessments) and inter-rater reliability (consistency among multiple evaluators) are prerequisites for valid judgements, confirming the robustness and trustworthiness of the evaluation process and the conclusions derived (Einhorn 1974, DeVellis and Thorpe 2022).

## 4. Demonstration Studies

### 4.1. Demonstration 1: Innovation in Software Application Updates

In a first demonstration study, we apply our systematic framework for innovation measurement to analyze and classify software application updates. Software application development is inherently generative, allowing for continuous product evolution through optimization, reuse, and novel combination of digital elements even after initial launch (Zittrain 2006). Digital platforms enhance this process by providing infrastructure and tools that reduce costs and accelerate development, fostering rapid experimentation and sequential innovation (Parker and Van Alstyne 2018). Low barriers to entry cultivate active developer communities and intense competition, leading to fast innovation cycles and early product releases (Lee and Raghu 2014, James et al. 2022). However, these same dynamics also incentivize the frequent release of low-quality updates or updates with little improvements to the software application and without introducing novel elements (Comino et al. 2019). When studying innovation behavior in many digital contexts, it is therefore crucial to differentiate between updates that merely refine existing products or "fix bugs", and those that contribute genuine innovation.

Recent methods for measuring innovation in software application updates include analyzing version numbers (Boudreau 2012, Tiwana 2015, Wen and Zhu 2019, James et al. 2022) and release notes (Foerderer et al. 2018, Leyden 2022, Kircher and Foerderer 2023). Relying on version numbers assumes that significant numerical changes indicate major updates, but developers often have incentives not to adhere to versioning protocol conventions[2], potentially distorting such metrics. Text analysis of release notes improves upon this issue but is mostly conducted using dictionary-based approaches or 'classical' machine learning methods. However, the vast heterogeneity in how developers describe updates and the absence of standardized terminology make it challenging to capture all variations accurately, leading to potential misclassifications.

---

[2] See https://semver.org/ for a popular example of such conventions.

Since datasets in such settings can be extremely large—often involving thousands or millions of descriptive texts—relying on human experts becomes impractical, albeit necessary, underscoring the need for advanced methods that can accurately assess innovation in software updates.

### 4.1.1 Data

An important aim of this demonstration study is to evaluate our proposed LLM framework and compare it to methods used in previous studies. To achieve this, we require a coded sample of software application updates, specifically requiring knowledge about the update and classification of its content. For comparability with previous studies, our data must include update information reported in release notes as well as the corresponding version numbers.

We obtained the data for this demonstration study from MixRank, a provider of app data. From a large dataset covering the complete update history of every app active on the Apple App Store between 2018 and 2021, we randomly sampled 4,000 update descriptions to form a pool for creating training and validation datasets. To ensure consistent analysis, we filtered the update descriptions to include only those written in English.[3] Each update description contains a corresponding release note and version number. On the App Store, these release notes are displayed in a section titled "What's New" and limited to 4,000 characters. Since 2018, it is mandatory for developers to specify update details and avoid generic texts.

To establish ground truth for the sample pool, a co-author and a research assistant coded the release notes in the pool according to a coding scheme developed by the research team through inductive coding. This scheme comprises seven content dimensions (classes) and was validated in interviews with software application developers (detailed in Section 4.1.2). A subset of n=1,000 update descriptions was independently coded by both raters to ensure high inter-rater reliability (observed agreement = 0.91, $\kappa$ = 0.87). Any discrepancies were discussed to be consistent across the remaining sample pool.

Subsequently, we generated two random subsamples from the labeled sample pool of n=4,000 release notes: a validation set of n=1,000 release notes serving as a holdout and a training set of n=2,000 release notes. Both subsamples were constructed to reflect a class distribution similar to that of the overall pool. The remaining n=1,000 release notes are used in further analyses to allow variations with training data sizes and class distributions.

---

[3] LLMs show increasing performance on multilingual datasets (e.g., OpenAI et al. 2023). Comparisons with alternative methods such as dictionary-based approaches are most reliable in English.

*4.1.2 Research Design*

We anchored our measurement objective in the literature's conceptualization of an innovative update in mobile software applications. In this context, innovative updates are defined as those that introduce additional, user-salient functionality or content to the app, rather than "under the hood" changes made primarily for internal purposes and imperceptible to consumers (Foerderer et al. 2018, Leyden 2022). In other words, the primary task of our prompt is to extract and analyze the content of release notes which we then use for an informed decision whether an update is innovative according to this definition.[4]

In a next step, we conduct inductive coding of release notes until reaching meaning saturation. This iterative process yielded six content dimensions that characterize the data structure of update release notes and enable us to distinguish between updates reflecting innovative activity (dimensions 1 and 2) and those intended primarily to fix bugs, introduce minor improvements, or promote the app (dimensions 3-7; cf. Figure 2).

----------------------------------------

Insert Figure 2 about here

----------------------------------------

To validate our measurement objective and data structure, we conducted several interviews with software application developers (Table 1). Following our proposed framework, both inform the prompt that maps the process of expert reasoning into an LLM-compatible format (cf. Figure 3). The validated content dimensions build the foundation for assigning human-coded labels to the release notes in our sample pool, including the training and validation datasets. Assigning these labels also serves as an additional check on data structure saturation, confirming that no further coding dimensions are necessary.

---

[4] We acknowledge that back-end modifications such as performance improvements, security hardening, or code refactoring may also be labelled "innovative" when viewed as process innovations. Our study, however, focuses explicitly on consumer-visible product innovations that add new functionality or content.

-----------------------------------------

Insert Table 1 and Figure 3 about here

-----------------------------------------

Our framework is applied to LLMs of various sizes, versions, and providers—both in their default state and after fine-tuning. For fine-tuning, the training data is prepared by integrating the release note text with its corresponding labeled output via our designed prompt. The generated outputs are evaluated against our validation dataset and benchmarked against the performance of methods from previous studies as well as 'classical' machine learning models, deep-learning approaches, and transformer-based pre-trained language models on the same validation dataset.

To replicate prior work, we incorporated the preprocessing steps, dictionaries and filters described in Agarwal & Kapoor (2023) and Kircher & Foerderer (2023) and we applied version logic similar to that used in Wen & Zhu (2019).[5] For benchmarking the other models, we trained each with the same labeled training data as our LLM fine-tuning, albeit without the prompts. To identify the best-performing model setup, we conducted several iterations with different embedding approaches and hyperparameter settings. More specifically, for each combination of embedding technique and classifier, we systematically optimized model hyperparameters via randomized search and stratified cross-validation across multiple dataset partitions. This entailed testing a range of embedding and model specific configurations to robustly identify optimal setups for each method (Goodfellow et al. 2016).

For instance, to optimize the Random Forest classifier used in this demonstration study, we constructed two pipelines that combined a TF-IDF vectorizer or precomputed GloVe word embeddings of varying dimensionality with the classifier. For the TF-IDF pipeline, we jointly tuned both vectorizer parameters (including the vocabulary size, n-gram range, minimum and maximum document frequency) and classifier hyperparameters (number of trees, maximum tree depth, minimum samples per split and leaf, feature subsampling ratio, criterion, and bootstrap usage) via randomized search and stratified cross-validation. In the GloVe setup, the text was transformed using different pre-trained GloVe models, and the resulting embedding vectors were used as direct input to the Random Forest. Here, we also systematically optimized the classifier's hyperparameters for each embedding variant. All search procedures were performed over 50 randomly sampled configurations, evaluated with 3-fold cross-validation, to robustly identify the best-performing model on the held-out validation data. The full hyperparameter grids, search procedures, and rationale for model selection are described in detail in our Online Appendix A.

---

[5] Kircher & Foerderer (2023) classify an update as innovative (feature update) if the release note contains at least one of the keywords "new", "added", "upgrade", or "major". Agarwal & Kapoor (2023) classify an update as innovative (new functionality) if the release note exceeds 200 characters or contains at least one of the keywords "introduce", "feature", "support", "performance", "improve", "enable", "update", "enhance", "modify", "optimize", "fast", "adjust", or "multitask". To account for differences in text representation and ensure consistency in the keyword matching process, we applied several preprocessing steps. We converted each release note to lowercase, tokenized it into individual words, and each word was reduced to its stem. The keywords in each dictionary were processed similarly. In Wen & Zhu (2019), the baseline treats every update release as an innovative update. A version-number rule is applied in robustness checks, classifying an update as innovative (major update) if the <major> field in the app's version number (first digit) changed. When replicating this approach, we additionally controlled for changes in the <minor> field in the app's version number (second digit).

*4.1.3 Results*

Table 2 reports performance (Accuracy, F1-score, Precision, Recall) and reliability metrics (Consistency Rate) for identifying innovative updates, comparing (1) rule-based and dictionary-based approaches from prior literature, (2) 'classical' machine learning (ML) models, (3) neural architectures such as convolutional neural networks (CNN), (4) earlier transformer-based pre-trained language models (e.g., ELECTRA, BERT), and (5) our framework applied to several large language models (LLMs) from different providers, with varied parameter sizes, and fine-tuning state. Accuracy measures the overall proportion of correct predictions, providing a straightforward assessment of model performance. However, to account for class imbalances and differentiate between false predictions, we use the F1-score, a standard composite metric calculated as the harmonic mean of Precision and Recall. Precision indicates the proportion of predicted innovative updates that are actually correct, while Recall measures the proportion of actual innovative updates correctly identified. Additionally, we evaluate model stability using the Consistency Rate (CR), which quantifies how reliably a model produces the same predictions across different runs.

----------------------------------------

Insert Table 2 about here

----------------------------------------

Relying on version numbers (i.e., distinguishing whether the "first digit" or the "second digit" of the version changed) alone yields low F1-scores (below 0.30 for the "Version First Digit" heuristic), confirming prior concerns that app developers do not consistently follow versioning protocols. Dictionary-based methods show better performance (F1-score up to 0.67), but remain noticeably behind more sophisticated approaches, reflecting the limited ability of fixed keyword lists to capture the richness of how new features are described in release notes.

Among 'classical' ML approaches (e.g., logistic regression, random forest, and support vector machines), F1-scores range from approximately 0.70 to 0.85, with logistic regression and support vector machine performing best within this category. The convolutional neural network performs similarly with an F1-score of 0.828, and early transformer-based pre-trained language models (PLM) perform even better with F1-scores close to 0.90. Crucially, all three categories of models can be run locally and reproduced under identical settings and training data, ensuring full replicability. However, each requires a substantial volume of labeled data to achieve their upper performance bounds (in this study, n = 2,000).

Even more compelling is the performance of several LLMs integrated into our framework. Without any additional training data (i.e., in their default configuration), and based on our framework, several LLMs achieve F1-scores on par with or better than PLMs, with GPT-4.1 reaching the highest F1-score of 0.901. Still, other default configurations such as GPT-4.1 Nano or Claude Haiku 3.5 fall below the performance of 'classical' ML models, suggesting that not all LLMs excel 'out of the box' and that models with fewer parameters perform worse than their larger counterparts. However, once fine-tuned with corresponding input-output pairs, performance improves across the board, with GPT-models edging F1-scores above 0.92.

Notably, even in their default configuration, LLMs maintain high precision, meaning they rarely mislabel non-innovative updates as innovative. In contrast, recall (i.e., the ability to correctly identify the full range of innovative updates) improves more markedly with fine-tuning. This suggests that, much like 'classical' ML models or earlier PLMs, additional training data helps LLMs to learn the varied and sometimes subtle language used to describe new features, thus better capturing the 'breadth' of innovative updates.

In practice, many LLMs are accessible only via an external API, making local deployment impractical or impossible. To nonetheless ensure reliability in these scenarios, we standardize our setup by setting the temperature parameter to 0 and by defining a model seed[6]. This adjustment produces deterministic outputs from the same model versions under identical prompts, with CRs above 0.98. The implications of variations in temperature settings, model selection, and their impact on both performance and reliability are further discussed in Section 5.

----------------------------------------

Insert Table 3 about here

----------------------------------------

Table 3 presents a more fine-grained, multi-class version of the update classification. Columns 1–7 reflect dimensions from the data structure and prompt, with per-class F1-scores, macro- and weighted-average results, and the CR across three independent runs. The macro F1-score averages performance equally across all classes, ensuring both common and rare updates are considered, whereas the weighted F1-score uses sample-based weights to reflect the dataset's true distribution. Comparing macro- and weighted-averages helps clarify whether models struggle on underrepresented classes (via macro-average) or show strong overall performance on the more frequent classes (via weighted-average). Large discrepancies between these averages can signal that certain classes are being overlooked.

Earlier dictionary- or version-based methods are not included in this analysis because they rely on fixed heuristics or keywords that cannot differentiate between subtle update types. 'Classical' ML and early PLMs (e.g., ELECTRA, RoBERTa) perform robustly overall but fail to detect rare classes when training data are scarce. Although misclassifying classes that serve a similar measurement objective may not drastically affect the overall innovation score, failing to detect a class altogether can skew outcomes and introduce bias.

By contrast, default LLMs demonstrate more balanced performance across all classes. When fine-tuned with the same training data used by ML or PLM approaches, LLMs (e.g., GPT-4.1 or GPT-4.1 Mini) can achieve a macro-average F1-score of up to 0.87, showcasing the advantages of our proposed framework and the capacity of novel LLMs to precisely classify content. Consistency rates (CR) for both default and fine-tuned LLMs remain very similar and consistently far exceed typical human inter-annotator agreement standards (Landis and Koch 1977, Carletta 1996, Artstein and Poesio 2008). Although the models are accessed via an API and not in a local setup, with the specified temperature and seed settings, both default and fine-tuned versions produce identical outputs across independent runs, with only rare discrepancies.

---

[6] Defining a seed parameter fixes the pseudo-random sampling state, so identical prompts and parameters return the same output when the model (*fingerprint*) is unchanged. Omitting the seed—even with temperature=0 calls—lets stochastic tie-breaks and sampling noise vary across runs and can reduce consistency across runs.

## 4.2. Demonstration 2: Originality in Product Reviews

In our second demonstration study, we apply and evaluate our framework to analyze and classify user generated product reviews. External sources of innovation such as customer and user ideas are invaluable for organizational innovation activities, providing access to knowledge and insights that may lie beyond organizational boundaries (Von Hippel 1978, Chesbrough 2003). However, research on crowdsourcing and idea evaluation highlights that identifying ideas that are both original and practically relevant, and that genuinely add value to organizations' innovation activities, remains challenging due to the overwhelming number of suggestions, their often unstructured nature, and evaluators' limited attention focus in idea contests and review platforms (e.g., Piezunka and Dahlander 2015).

Previous methods that analyze customer ideas and needs in product reviews often rely on expert committees or advanced NLP techniques backed by large training datasets (e.g., Timoshenko and Hauser 2019, Zhang et al. 2021). While effective, these approaches can easily become resource-intensive, requiring extensive training data, specialized NLP expertise, and significant involvement from many domain experts. We illustrate LLMs potential to generate valid evaluations of user ideas with less reliance on extensive data or expert involvement.

### 4.2.1 Data

Our dataset, sampling, and preprocessing procedures are comparable to those used in Demonstration Study I. From a large dataset covering the complete review history of all apps that have been listed at least once in the top-100 overall or category-specific rankings, we randomly sampled 4,000 reviews written in English.

To establish ground truth for the sample pool, a co-author and a research assistant independently coded the reviews using a coding scheme developed through inductive coding by our team. This scheme comprises nine content dimensions (detailed in Section 4.2.2). A subset of n=1,000 review descriptions was independently coded by both raters to assess inter-rater reliability (observed agreement = 0.89, macro-average $\kappa$ = 0.77), and any discrepancies were discussed.

In line with Demonstration Study I, we generated two distinct random subsamples from the labeled pool of n=4,000 reviews: a validation set of n=1,000 reviews serving as a holdout, and a training set of n=2,000 reviews. Both subsamples were constructed to reflect a representative class distribution of the overall pool. The remaining n=1,000 reviews of the labeled pool serves to back additional analyses to explore variations in sample sizes and class distributions.

*4.2.2 Research Design*

Our measurement objective builds on the premise that product reviews often contain valuable suggestions, ideas, and expressions of customer needs (Timoshenko and Hauser 2019, Zhang et al. 2021). These reviews are multifaceted, ranging from brief comments to extensive narratives where users may suggest or report several issues at once. As such, our prompt is aimed at distinguishing between different types of content and to identify instances where multiple topics are addressed simultaneously.

----------------------------------------

Insert Figure 3 about here

----------------------------------------

To further inform prompt design, we conducted inductive coding of reviews until meaning saturation was reached. As shown in Figure 3, this process resulted in nine content dimensions that characterize the reviews' data structure and provide a means to differentiate them at a highly granular level. Both our measurement objective and the data structure are then integrated in the final design of our prompt (see Figure 4) and the content dimensions serve as the foundation for labeling reviews in our sample pool.

----------------------------------------

Insert Figure 4 about here

----------------------------------------

Our framework is applied to the same LLMs used in Demonstration Study 1. To prepare training data for fine-tuning, we created input-output pairs by combining every merged prompt and review text with its corresponding manual label. The LLM generated outputs are then evaluated against our validation dataset and benchmarked against the performance of 'classical' machine learning models, deep-learning approaches, and transformer-based pre-trained language models on the same validation dataset.

For benchmarking against the other models, we used the same labeled training data as our LLM fine-tuning, though without the prompt. Given the multi-label nature of our task, we employed a one-vs-rest strategy, where each label is encoded and classified as an independent binary feature to capture the presence of multiple labels per review. For instance, if a review's labels were 8 and 4, this method converted it into a binary vector [0 0 0 0 1 0 0 0 1] representing the presence of labels from "0" through "8". Similar to Demonstration Study 1, we experimented with different embedding techniques and hyperparameter settings to identify the best-performing setup. The rationale behind our configuration for each model is detailed in our Online Appendix A.

Although product reviews, and particularly app reviews, are a common subject in NLP tasks and have been extensively explored for feature extraction and content classification (Khlifi et al. 2024, Motger et al. 2024, e.g., Gunathilaka and de Silva 2025), no study has, to our knowledge, tested such fine-grained content differentiation or demonstrated LLMs' capabilities in a multi-label setting.

*4.2.3 Results*

Table 4 shows the results of applying our framework to an analysis of reviews, reporting comparable performance and reliability metrics as in Demonstration Study 1. The nine columns (1–9) correspond to the content dimensions derived from the data structure and included prompt. Because each review can receive multiple labels in the one-vs-rest approach, the total number of labeled instances (n) exceeds 1,000, even though the holdout sample contains 1,000 reviews.

----------------------------------------
Insert Table 4 about here
----------------------------------------

The overall observations from Demonstration Study 1 are corroborated here. 'Classical' machine learning models, with random forest showing the highest Macro F1 score (0.401), and K-Nearest-Neighbor (Macro F1 score = 0.208) yielding the lowest performance. Applying pretrained language models (PLMs) results in modest improvements, with XLNet achieving a Macro F1 score of 0.436. Notably, applying our framework in LLMs outperforms these approaches, with even default configurations achieving superior results. The best performance is obtained through fine-tuning, with GPT models achieving top Macro F1 scores (e.g., GPT-4.1 = 0.706).

Models that leverage additional training data or benefit from fine-tuning particularly excel in classes with larger sample sizes. Interestingly, default versions of LLMs can achieve F1 scores comparable to their fine-tuned counterparts and deliver moderate performance in scenarios where traditional methods struggle, such as with underrepresented samples or linguistically complex cases (e.g., class 6). Further performance enhancements are achievable by fine-tuning labeled prompt-review-output combinations.

Consistent with Demonstration Study 1, the temperature parameter is set at 0, and the model seed is fixed to systematically assess reliability metrics. Also in this multi-label setting, outputs across three independent runs remain highly consistent, with consistency rates persistently above 0.94. A detailed analysis of the impact of varying temperature settings on model performance and reliability is presented in Section 5.

## 5.  Implications of Design Decisions

Design decisions, such as model selection, prompt engineering, temperature settings, training data distribution, and data size, are critical when applying the proposed evaluation framework because they can affect the model's performance, output consistency, and applicability to specific tasks. These decisions shape how well the model aligns with the intended use case and determine the quality, reliability, and interpretability of its outputs (Cheng et al. 2024, Carlson and Burbano 2025). Therefore, transparency in design decisions is demanded. For reviewers and editors tasked with verifying approaches, understanding these decisions is essential to critically assess the validity and reproducibility of results, as well as to ensure that the chosen decisions are appropriate for the research objectives.

*5.1. Model Selection*

When applying the proposed evaluation framework, researchers have to select an appropriate LLM from the numerous available models from providers like OpenAI (GPT), Anthropic (Claude), and Mistral (Ministral/Mistral). Their models differ in attributes such as model age, model size in terms of parameters and pre-training data scale, reasoning capabilities, and general evaluation benchmarks[7]. In both demonstration studies, we initially included the most recent models from these providers and have now extended our analyses to cover all currently available models from each provider.

-----------------------------------------

Insert Table 5 about here

-----------------------------------------

Table 5 displays the performance results by model provider for both demonstration studies. The results confirm the earlier observations from Tables 3 and 4 that larger models outperform their smaller counterparts in default configurations. For example, GPT-4o demonstrated notably higher performance (Macro F1-Scores of (1) 0.602 and (2) 0.640) than GPT-4o-mini (Macro F1-Scores of (1) 0.484 and (2) 0.576) or the predecessor GPT-3.5 Turbo (Macro F1-Scores of (1) 0.439 and (2) 0.564). This pattern holds consistently across Claude and Mistral model families. However, model recency does not guarantee superior performance within similar size categories. In Demonstration Study 2, Claude Sonnet versions showed comparable results of Sonnet 3.5 (Macro F1-score = 0.649), Sonnet 3.7 (Macro F1-score = 0.638), and Sonnet 4 (Macro F1-score = 0.631), while in Demonstration Study 1 earlier versions even outperformed the newest model, with Sonnet 3.5 achieving a Macro F1-score of 0.644 and Sonnet 3.7 achieving 0.613, compared to Sonnet 4's score of 0.588.

---

[7] General-purpose evaluation of language models relies on standardized benchmarks such as GPQA, Humanity's Last Exam, and Massive Multitask Language Understanding (MMLU). These tests probe a wide range of competencies through multiple-choice and short-answer questions that span dozens of academic and professional subjects. Each item is written by domain experts and linked to a single, unambiguous, and easily verifiable answer that cannot be retrieved through a quick web search, ensuring that scores reflect not a simple information lookup. Complementing these broad assessments are specialized benchmarks that focus on domain-specific abilities—for example, mathematics (e.g., GSM-8K), programming (e.g., HumanEval), or creative writing. They emphasize complex problem-solving within their respective fields. Thought experiments such as the Turing Test add a qualitative dimension by examining how convincingly a model can mimic human-like understanding and conversation. For fair comparison, models are evaluated in their default configurations. Past developments have shown that newer and larger-sized model versions or explicitly reasoning-optimized models tend to achieve the highest scores.

Fine-tuning significantly improves model performance across all tested models. Most remarkably, fine-tuned GPT-3.5 models exceeded the performance of GPT-4.1 models in Demonstration Study 1 and nearly matched their performance levels in Demonstration Study 2. Smaller Mistral models (Ministral and Open Mistral Nemo), designed primarily for local implementation with minimal hardware requirements, initially performed poorly with F1-scores below 0.4 in Demonstration Study 1. However, these models showed substantial performance improvements through fine-tuning, approaching the performance levels of larger fine-tuned Mistral model classes.

The performance of new reasoning models varies across providers. While Claude Opus models underperformed compared to Sonnet models in Demonstration Study 1 and matched them in Demonstration Study 2, GPT-o3 achieved performance values comparable to the best fine-tuned GPT models. This capability simplifies large-scale text analysis implementation within our framework without requiring additional training data. However, reasoning models are considerably more resource-intensive, requiring up to several hundred times more tokens to generate outputs similar to non-reasoning models.

Taken together, larger-sized or reasoning models already achieve strong results in their default versions. Provider differences are minor, with GPT-based models slightly outperforming others particularly when fine-tuned. Notably, when fine-tuned on adequately sized, task-specific datasets, mid- and large-sized models exhibited negligible performance differences regardless of provider or size. Consequently, researchers applying our framework should systematically monitor multiple LLM candidates to identify the best fit for their evaluation tasks rather than assuming that higher parameter count, more recent releases, or provider reputation automatically translates into superior outcomes.

In addition to performance considerations, practical factors such as usage costs, energy consumption and model accessibility are also worth considering. The majority of the models utilized in our demonstration studies are proprietary, accessed via APIs with usage-based fees, typically lower for smaller-sized models. Many LLM candidates are also available as open-source, downloadable, and runnable on local computing infrastructure. However, while smaller models such as Ministral 8B or Open Mistral Nemo are compatible with standard computational resources, mid-sized open-source models like Mistral Small already require significantly enhanced computational capabilities, especially when fine-tuning them.[8] Due to such resource constraints, our demonstration studies did not include locally run tests of models comparable in size to those analyzed via APIs. Still, we expect their performance to align with our observed results from proprietary models, particularly since models such as Mistral's offerings are available in both proprietary and open-source versions.

## 5.2. Prompt Engineering

Prompt engineering refers to the strategic modification of prompts to enhance the performance of LLMs on specific tasks (Brown et al. 2020). Studies in several contexts demonstrate that these modifications can yield significant performance improvements for LLM applications (see, for example, the systematic survey by Sahoo et al., 2024). In both demonstration studies, we assessed different prompt engineering techniques using the GPT models: few-shot prompting, automatic chain-of-thought, manual chain-of-thought, and contrastive chain-of-thought. Few-shot prompting involves presenting the model with a limited number of example input-output pairs (Brown et al. 2020). Automatic chain-of-thought prompting aims at generating intermediate reasoning steps by adding a "Let's think step by step" instruction at the end of the prompt (Zhang et al. 2022). In contrast, manual chain-of-thought-prompting integrates explicitly formulated logical reasoning chain to generate an output into the prompt (Wei et al. 2022). Contrastive chain-of-thought-prompting adds learning from mistakes by providing both valid and invalid reasoning explanations alongside the original prompt (Chia et al. 2023). The prompt texts for each technique are documented in Appendices A1 and A2.

-----------------------------------------

Insert Table 6 about here

-----------------------------------------

Table 6 displays the performance results by prompt engineering technique for both demonstration studies. For default models, few-shot prompting showed slight performance

---

improvements compared to the "default" prompt. However, it is important to understand that few-shot prompting introduces the risk of selection bias stemming from the choice of examples. Chain-of-thought prompting yielded mixed results. While GPT-4.1 Mini and GPT-4.1 performance decreased with automatic and manual variants but improved with contrastive chain-of-thought prompting in Study 1, all chain-of-thought variations slightly improved performance in Study 2. Notably, GPT-4.1 Nano performance dropped strongly with chain-of-thought prompting in Study 1 (baseline Macro F1-Score with the default prompt = 0.440).

Interestingly, prompt engineering had often only minimal positive performance implications on fine-tuned models, despite these models originally being fine-tuned with another prompt. This suggests that prompt modifications can still benefit performance despite substantial prompt-specific training. However, the fine-tuned GPT-4.1 Nano model showed a different pattern, with performance declining when prompts deviated from training data. In Study 1, the "default" prompt achieved a Macro F1-score of 0.710, while alternative prompt variations few-shot (0.650), automatic chain-of-thought (0.593), manual chain-of-thought (0.554), and contrastive chain-of-thought (0.649) scored lower. Similar declines, though less pronounced, appeared in Study 2.

The results highlight that prompt engineering can impact performance across different models and contexts, even for fine-tuned systems. However, for larger models we observe when prompts are carefully crafted and validated with expert input—particularly by guiding through data structures and classification patterns—the potential for performance gains through prompt engineering appear limited.

*5.3. Temperature Value*

The temperature is a hyperparameter used in language models to control how random or predictable the generated outputs are (Ackley et al. 1985). When models produce outputs, they first compute raw scores (logits) for each possible next token. These logits are converted into probabilities using the softmax function, which normalizes them so that they sum to 1. The temperature parameter influences the shape of this probability distribution by redistributing the probabilities across tokens. A higher temperature ($> 1$) flattens the distribution, increasing uncertainty and randomness. Conversely, a lower temperature (< 1) sharpens the distribution, emphasizing tokens with higher probability and reducing randomness.[9] Temperature values typically range from 0 to 2, though some providers use different scales (e.g., Claude models cap at 1). Setting temperature = 0 removes randomness entirely, causing the model to always choose the token with the highest probability (greedy sampling). Conversely, higher temperature settings can introduce randomness, enabling more creative or unexpected outputs (Peeperkorn et al. 2024). However, increased randomness comes at the expense of output reliability across repeated runs, which is a critical factor in research-oriented evaluations.

We analyze the implications of this setting in our demonstration studies by gradually increasing temperature values. Figure 6 displays F1-scores of the first run and consistency rate across three independent runs for temperature values of 0.5, 1.0, and 1.5 for both demonstration studies. The metrics indicate that higher temperature settings do not improve model performance and instead result in a slight decline in performance alongside a reduction in reliability as measured by the consistency rate for some models. Most changes are minor with GPT models in particular showing stable performance and reliability. For example, in Demonstration Study 1, our framework applied to a GPT-4.1 (default) model shows a decline in macro averaged F1-score from 0.657 at temperature = 0.0 to 0.638 at 0.5, 0.636 at 1.0, and 0.650 at 1.5. The consistency rate changes from 0.970 at temperature = 0.0 to 0.98 at 0.5, 0.979 at 1.0, and 0.968 at 1.5. In contrast, a more pronounced decrease in consistency rate is evident for the Claude models and Mistral Large (default). For instance, the consistency rate of Mistral Large falls from 0.973 at temperature = 0.0 to 0.828 at temperature = 1.5 in Demonstration Study 1, and from 0.911 to 0.611 in Demonstration Study 2.

----------------------------------------

Insert Figure 6 about here

----------------------------------------

Practically, this decrease in consistency means that at higher temperature settings, models like Mistral Large and Claude are less likely to generate the same output for identical input across runs, which undermines reproducibility. When operationalizing text-based variables this variability can introduce noise, making it difficult to reliably measure constructs, replicate results, or compare findings across studies. In addition, increased randomness raises the effort required for output cleaning, as deviations from the defined format become more frequent. This problem is particularly pronounced at the maximum temperature setting (2.0), where our tests revealed that model outputs often deviated so strongly from expected formats that label identification became unfeasible, ultimately rendering these outputs unusable for analysis and leading to their exclusion from further evaluation. Based on these findings, we recommend that researchers adopt a conservative default temperature ($\leq 0.5$) for tasks requiring strict output structure or having high reproducibility demands, unless a higher value is conceptually justified. Our exact procedure for label identification in noisy LLM outputs is detailed in the Online Appendix B2.

### 5.4. Training Data Distribution and Data Size

As demonstrated in previous analyses across both demonstration studies, LLMs fine-tuned with additional training data can lead to substantial improvements in performance. Likewise, earlier ML and PLM approaches have performed well in both studies. However, when the distribution of the training data reflects that of the real-world holdout sample, some labels become underrepresented. Models that rely heavily on these training data distributions tend to perform worse on underrepresented labels compared to those labels that have sufficient representation in the training set.

To address this imbalance, two primary approaches can be considered (e.g., Johnson and Khoshgoftaar 2019). The first approach focuses on the algorithmic level, where model weights are adjusted to assign greater importance to underrepresented classes, thereby imposing heavier penalties for classification errors related to these classes in the training or fine-tuning process more heavily. Essentially, this method instructs the model to pay greater attention to rare labels by amplifying the consequences of mistakes associated with them. The second is at the data-level, adjusting the training data directly. This involves modifying the distribution of the training set through strategic oversampling of underrepresented labels, thereby enhancing model exposure to these less frequent examples. This method is particularly relevant and effective when algorithm-level adjustments are infeasible, as is often the case with fine-tuning proprietary LLMs.

Building on these considerations, it is important to recognize that, in practice, particularly data-level approaches depend on the availability of sufficiently large datasets. In previous analyses, we deliberately created extensive training and validation datasets to avoid performance differences arising solely from data limitations. However, creating large training and validation datasets is often time-consuming, complex, or constrained by data or the availability of human coders, depending on the evaluation task. To better understand how data constraints interact with distributional characteristics, we systematically reduced the size of the training datasets for both balanced and representative class distributions. Specifically, we generated subsamples with n = 1,000, n = 500, n = 250, and n = 100, maintaining either a balanced or representative class distribution as required.

-----------------------------------------

Insert Figure 7 about here

-----------------------------------------

Figure 7 presents the macro-averaged F1-scores as training data size increases, for both balanced and representative class distributions across both demonstration studies. In Demonstration Study 1 with the representative (real-world) class distribution, clear differences between models emerge. The fine-tuned GPT-4.1 Nano and Mini models outperform their default counterparts even with smaller sample sizes, showing steady gains up to n = 500, a slight decline at n = 1,000, and reaching peak performance at n = 2,000. The larger GPT-4.1 model also starts above its default version with small sample sizes, dips below default at n = 500, but then shows a steady increase, plateauing from n = 1,000 onwards. In contrast, both fine-tuned Mistral models initially underperform relative to their default versions, only surpassing default performance at the largest sample size (n = 2,000). When using a balanced class distribution, performance increases more smoothly for all models as sample size grows, and maximum F1-scores are higher than with the representative distribution. Still, the Mistral Small model does not surpass its default until n = 1,000, and Mistral Large until n = 2,000. In Demonstration Study 2, balanced distributions again yield steady growth and higher peak F1-scores compared to representative distributions, and no fine-tuned model underperforms its default version at any sample size. The representative distribution in study 2 shows less pronounced overall gains, echoing the overall moderate improvements seen in study 1.

Taken together, larger-sized or reasoning models already achieve strong results in their default versions. Provider differences are minor, with GPT-based models slightly outperforming others particularly when fine-tuned. Notably, when fine-tuned on adequately sized, task-specific datasets, mid- and large-sized models exhibited negligible performance differences regardless of provider or size. Consequently, researchers applying our framework should systematically monitor multiple LLM candidates to identify the best fit for their evaluation tasks rather than assuming that higher parameter count, more recent releases, or provider reputation automatically translates into superior outcomes.

In addition to performance considerations, practical factors such as usage costs, energy consumption and model accessibility are also worth considering. The majority of the models utilized in our demonstration studies are proprietary, accessed via APIs with usage-based fees, typically lower for smaller-sized models. The tested Mistral models are also available in open-weight versions, and can be run on local computing infrastructure. However, while smaller models such as Ministral 8B or Open Mistral Nemo can operate effectively on standard computational resources, mid-sized open-weight models like Mistral Small already require significantly enhanced computational power, especially when fine-tuning them. Due to such resource constraints, our demonstration studies did not include local testing of models comparable in size to those analyzed via APIs. Still, we expect their performance to align with our observed results from proprietary models, particularly given models such as Mistral's offerings are available in both proprietary and open-weight versions. Since their architectures are identical, any performance discrepancies would likely stem from variations in computing infrastructure.

## 6. Discussion

### 6.1. Contributions to Innovation Research

The measurement of innovation-related phenomena is often constrained by the challenges inherent in identifying and assessing contextual information (Boudreau et al. 2016). In many cases, such evaluation relies on expertise-based judgment, yet access to qualified experts is often limited. Moreover, analyzing large datasets is impeded by the inherent cognitive limits of human evaluators. As the volume of the data increases, the cognitive load on experts can lead to greater reliance on heuristics and biases, which may impair their ability to evaluate effectively (Tversky and Kahneman 1974). Empirical research in innovation has therefore mainly been conducted in contexts where reliable measures are readily available. Our research addresses these challenges by developing an LLM-based framework. Our framework is based on the premise that certain contextual cues are highly indicative of expert judgement (Einhorn 1974). We apply qualitative methods and collaborate with experts to identify these cues and assess their relevance and integrate these insights systematically into prompt design and fine-tuning process for the LLM. A key strength of our framework is that it reduces the reliance on human experts while still achieving comparable evaluations. The scalability enabled by this approach simplifies empirical innovation research across a broader range of contexts, and also provides opportunities for additional robustness checks and analytical perspectives.

We successfully illustrate the applicability and utility of our framework in two demonstration studies conducted in different innovation contexts. Although the validation exercises are specific to innovation in digital platform ecosystems, they capture two central concepts in innovation research: the outputs of innovative activity and the generation of ideas, holding promise for extension to other innovation contexts as well. However, even the best performing models in our demonstration studies still misclassify roughly one in ten instances (precision $\approx .90$). Any high-stakes deployment should retain a human-in-the-loop review and undertake a careful error-impact analysis tailored to the application context.

Furthermore, the application of our framework makes specific contributions to the research areas of these two demonstration studies by enhancing existing metrics. First, in studying innovation in software application development in digital platform ecosystems (Wen and Zhu 2019, e.g., Agarwal and Kapoor 2023, Kircher and Foerderer 2023), we improve measurement performance, assessed by F1-scores, of the innovation variable by up to 71% relative to previous studies (see Table 2), and additionally enable a more fine-grained differentiation of update content. Second, in measuring the innovativeness of user-generated ideas (e.g., Timoshenko and Hauser 2019, Zhang et al. 2021), our framework allows for a more detailed, multi-class screening of ideas.

The implications of these improvements are twofold. First, they introduce new levels of analysis and promising directions for future research. Specifically, the capability to examine update release notes and product reviews at such detailed levels allows researchers to accurately trace how user innovations are adopted and integrated over time. Consequently, researchers can gain deeper insights into when and how user-suggested ideas become part of products and assess the extent to which these incorporations contribute to outcomes like sustained competitive advantage. For instance, products with larger and more engaged user communities may disproportionately benefit from a greater influx of user-generated ideas, potentially accelerating innovation cycles or enhancing adaptability to market demands. Moreover, such detailed analyses facilitate better evaluation of the impact of external events that influence development and user behavior. For example, regulators and platform owners of digital platforms are increasingly emphasizing user privacy by introducing privacy-persevering regulations. The refined metrics demonstrated in our studies allow researchers to quantify the effects of these regulations, such as their influence on application development practices and user perceptions regarding the regulations and corresponding implementations by developers. Second, the improvements may increase the validity of empirical findings and conclusions such as in studies on complementor innovation on app platforms, by avoiding potential systematic measurement errors that could result in substantial bias in coefficient estimates (Millimet and Parmeter 2022, e.g., deHaan et al. 2024). Future research, through replication studies and further exploration, is needed to fully assess the implications of these enhancements.

Finally, an essential component of our framework is assessing the impact of design decisions made by researchers on performance and reliability (Cheng et al. 2024, Carlson and Burbano 2025). Our demonstrations show that variations in these design decisions can affect outcomes, highlighting the importance of detailed reporting for reproducibility when applying LLMs in research. This insight is particularly valuable for studies examining the implications of AI on innovation, where rigorous experimental design and transparency are essential for the credibility of results.

*6.2. The Role of the Expert Evaluator*

Our framework is centered at relevant domain expertise being available for cue identification and weighting. The underlying assumption is that bounded rationality and imperfect judgement in evaluation tasks affects all possible evaluators, while those whose expertise aligns most closely with the evaluation context exhibit lower uncertainty in their judgements (Csaszar and Eggers 2013, Boudreau et al. 2016, Böttcher and Klingebiel 2025). This is because domain experts are able to disproportionately identify and interpret more relevant information compared to less specialized evaluators (Kahneman et al. 1982, Johnson et al. 1982, Ericsson and Smith 1991).

However, we also recognize that identifying the most suitable expert *ex ante* is challenging. There exists a risk of cognitive entrenchment, where experts may be less inclined to identify and integrate new information that challenges their existing views (Dane 2010, Furr et al. 2012, Almandoz and Tilcsik 2016). As a result, they may inadvertently extrapolate from their established knowledge into domains where their expertise is less applicable. This misalignment can degrade evaluation performance, even to a point that their judgment becomes inferior to those of generalist evaluators (Csaszar and Eggers 2013, Böttcher and Klingebiel 2025).

Although the framework is designed to enforce robust evaluation processes, it is not immune to the risk of replicating or even reinforcing such evaluator bias. These risks are partially mitigated by the framework's data-centric approach, where the extraction of the data structure is carried out by the researcher, and domain experts are engaged primarily for validation and adjustment purposes. Still, there remains a potential for bias, particularly when fine-tuning with biased training data is involved not carefully managed.

Given these limitations, alternative evaluation scenarios are worth considering. In contexts where identifying the most relevant domain experts is challenging, integrating insights for instance, from crowds (e.g., Mollick and Nanda 2016, Wimbauer et al. 2019) or evaluator panels (e.g., Criscuolo et al. 2017) may yield an increase in evaluation performance. The primary advantage of the proposed framework lies in its data-centric approach, where expert validation is confined to the data structure. This design makes the framework agnostic to the type and format of expert input. Future research could explore how integrating different evaluator groups and formats (e.g., ranking, voting, aggregation) in our framework impacts performance, providing deeper insights into the trade-offs and benefits of each approach. This exploration is particularly relevant as evaluation methods for LLMs continue to evolve, with hybrid strategies combining human and automated evaluations emerging as a promising direction.

*6.3. Ethical and Practical Considerations*

LLMs are often considered "black boxes" due to their complexity and opacity. They are pre-trained on exceptionally large datasets and build on advanced deep learning architectures with billions of parameters. Despite their impressive performance, our current understanding of their internal mechanisms is limited (Bommasani et al. 2021), and independently constructing comparable models at similar performance scales is typically beyond the capabilities of individual researchers or smaller institutions. Moreover, the processes by which LLMs generate outputs and the biases introduced during pre-training, particularly with datasets undisclosed, are not yet fully understood. Variations between different models, or even between different versions of the same model, are both inevitable and expected. When applying the framework, it is therefore essential to systematically validate the outputs against a holdout sample or established ground truth to accurately assess performance and detect potential systematic measurement errors.

Practically, using LLMs involves navigating operational considerations related to both proprietary and open-source models. Proprietary models, typically accessed via APIs, often provide minimal and non-specific information about their internal configurations and are subject to potential modifications or discontinuation by providers. Therefore, detailed documentation of the exact model version used and transparent reporting of all design decisions is essential for ensuring reproducibility. While proprietary models still offer advantages in efficiency and effectiveness, API costs, despite decreasing over time[10], may accumulate significantly when working with large datasets. Open-source models, on the other hand, reduce the risks associated with proprietary models, such as unexpected accessibility changes or alterations in performance. However, using medium or large-sized open-source models locally demands substantial hardware infrastructure. Alternatively, cloud-based deployment can also incur significant scaling costs. Despite platforms like Hugging Face considerably reducing entry barriers, substantial technical expertise remains essential.

---

[10] LLM providers iterate frequently on model capabilities and pricing, so usage fees can drop sharply within a few months. For example, OpenAI's GPT-4o model launched in May 2024 at US $5 per Million input tokens and US $15 per Million output tokens (model ID *gpt-4o-2024-05-13*), but the August 2024 revision (*gpt-4o-2024-08-06*) cut those rates to US $2.5 / $10. Likewise, the o3 reasoning model (*o3-2025-04-16*) fell from US $10 / $40 at release in April 2025 to US $2 / $8 in June 2025 (see https://platform.openai.com/docs/pricing for further details).

Given the rapid advancements in LLM technology, questions may arise regarding the long-term relevance of our proposed framework. The proposed framework is designed to be model-agnostic, allowing it to leverage ongoing advancements in LLM capabilities and adapt to future improvements in the field. By focusing on mapping expert cues into the evaluation process, the framework provides a robust and adaptable foundation for scaling the assessment of innovativeness in dynamic and diverse scenarios.

**Figure 1 Framework for mapping expert insights into LLM applications**

| Expert Evaluation | LLM Process Mapping | | Design Decisions |
|---|---|---|---|
| *Task Definition* | Prompt Section 1 | - Researcher describes task and defines desired output according to the measurement objective.<br>- Task description and output definition are included in the prompt. | Researcher can apply prompt engineering to optimize model performance |
| *Data Engagement* | Inductive Coding AND Prompt Section 2 | - Researcher conducts inductive three-pronged (open, axial, selective) coding procedure with the data until code/thematic saturation. is reached.<br>- Data structure is validated with expert(s) to assure meaning saturation and adequate structuring of cues.<br>- Data dimensions (or categories) are included in the prompt. | |
| *Relate to Prior Knowledge* | Prompt Section 3 OR Manual Output Interpretation | - Expert assesses relevance and weight of each data dimension (or category) for the measurement.<br>- Relevance and weights are included in the prompt or researcher interprets model output based on the relevance of dimensions for the defined task | |
| *Judgement Formation* | Output validation AND Fine-Tuning | - Researcher and expert validate model output and measurement<br>- Researcher labels training and validation data for model fine-tuning on desired outcome (input: prompt + text data, output: classification/ measure in expected format) | General:<br>- Model selection<br>- Temperature setting<br><br>Fine-Tuning:<br>- Training data size<br>- Data distribution |
| *Reliability Assurance* | Design Decisions | - Researcher analyzes implications of model selection, prompt engineering, training data, and temperature setting on consistency rate of repeated runs<br>- Researcher optimizes design decisions for reliability and documents them | |

**Figure 2 Data structure for software application updates (Study 1)**

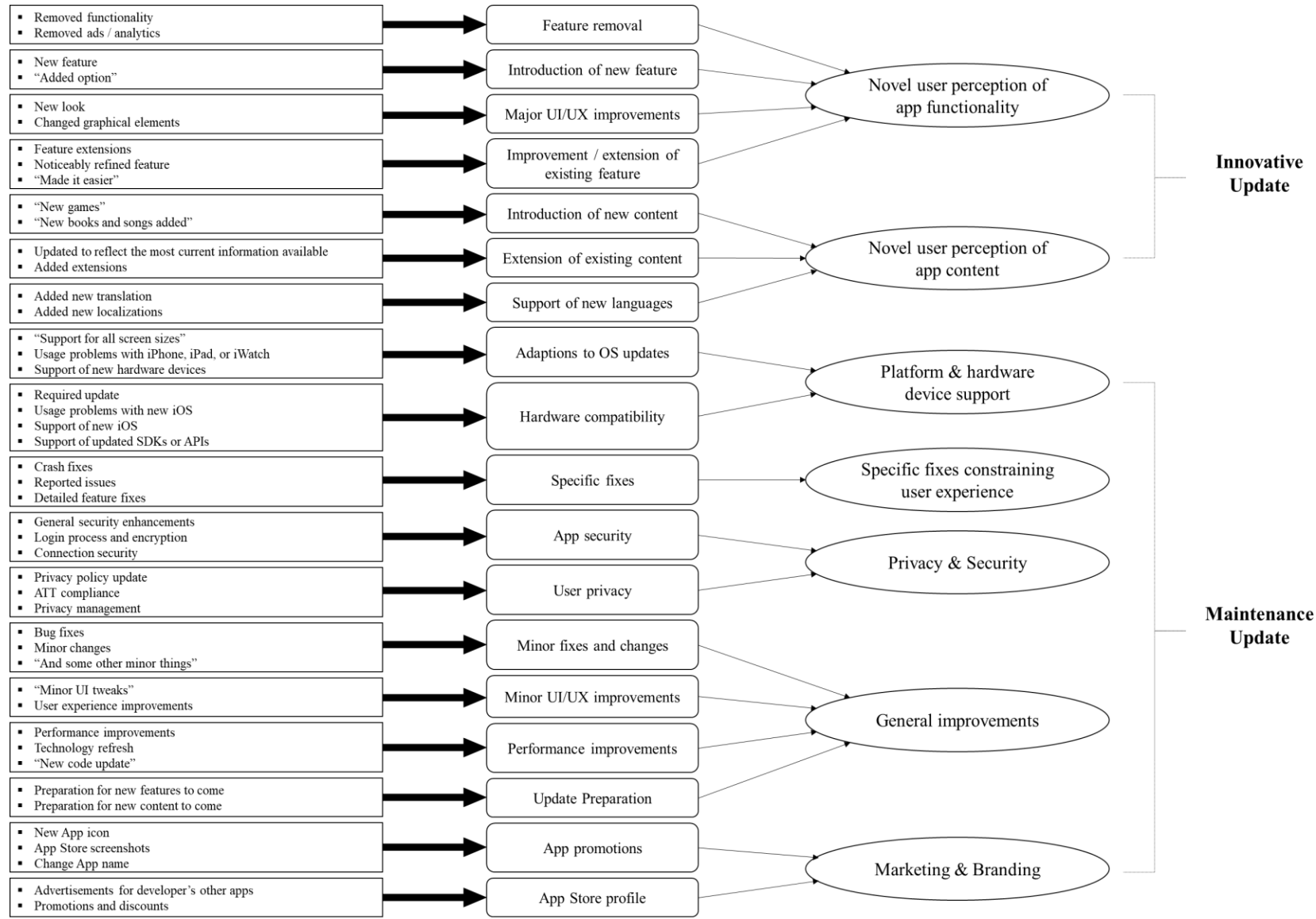

**Figure 3 System prompt and prompt breakdown for software application update classification (Study 1)**

| | |
|---|---|
| ONLY provide the category number (1-7) in response. Determine the category for the following app update text. If multiple categories seem applicable, always choose the lowest category number (1<2<3<4<5<6<7): | **Section 1**<br>Output definition and task description according to measurement objective |
| (1) Novel features - Introducing or enhancing significant functionalities that modify user experience<br>(2) Content extensions – Adding or expanding curated or user-facing content<br>(3) Platform and device support – Enabling compatibility with new OS versions, development kits, device types, or hardware features<br>(4) Specific fixes of bugs - Addressing distinct known issues<br>(5) Privacy & Security - Strengthening user data protection, permissions, or security protocols<br>(6) General Improvements – Tweaks addressing undisclosed bug fixes, performance improvements, minor changes, or UI/UX polish<br>(7) Marketing & Branding – Changes purely about promotional or visual branding assets | **Section 2**<br>Expert validated data structure |
| App Update: [*RELEASE NOTE*] | **Section 3**<br>Text data for evaluation |

**Figure 4 Data structure for product reviews**

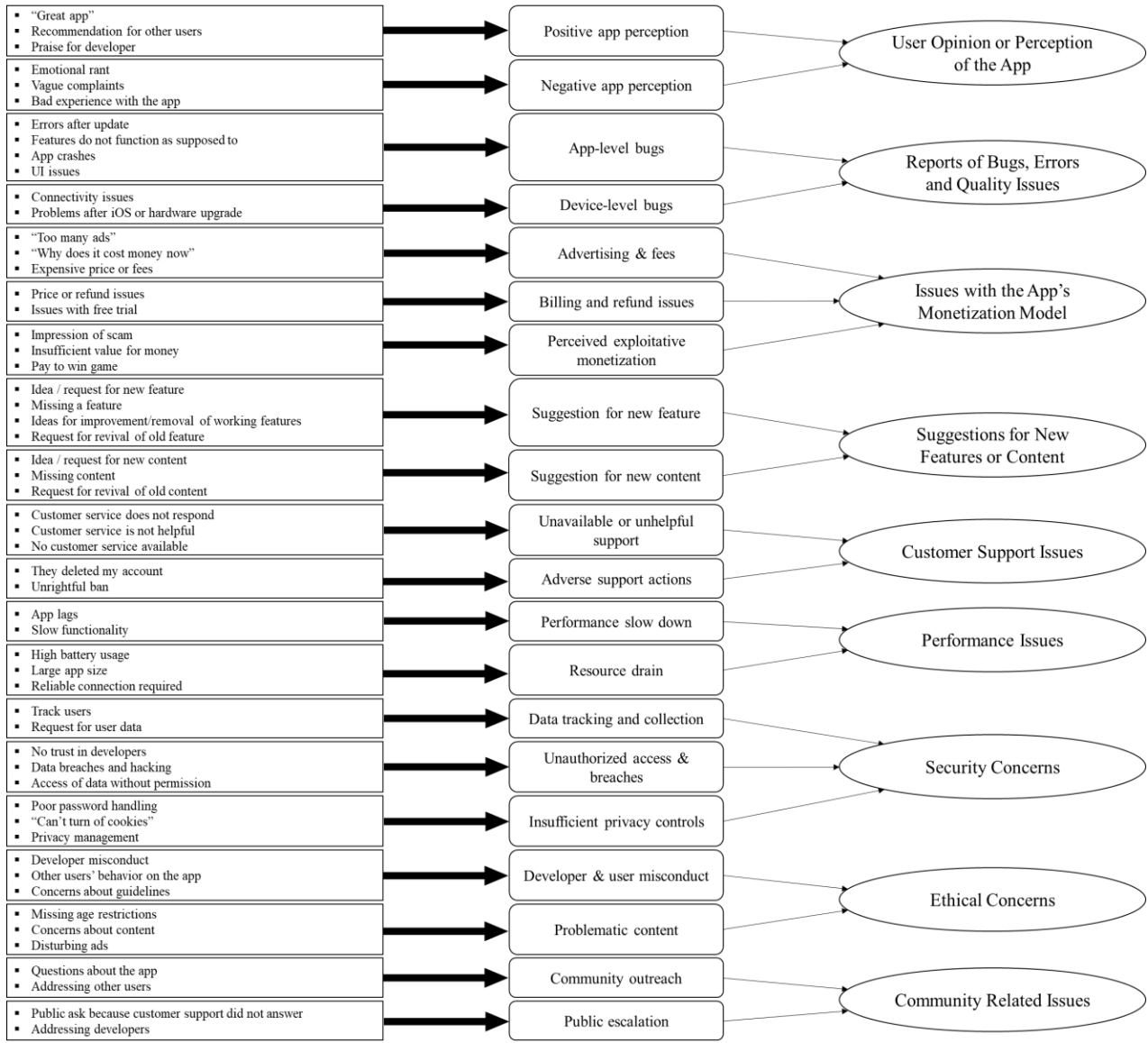

**Figure 5 System prompt and prompt breakdown for product review classification (Study 2)**

| | |
|---|---|
| ONLY provide a number (0-8) in response. Categorize the following app review text by assigning the most fitting category/categories out of the following nine categories.<br>If the text contains elements from multiple categories, provide the categories separated by ; | **Section 1**<br>Output definition and task description according to measurement objective |
| (0) User Opinion without specific reports, issues, suggestions - i.e., review only about good/bad perception of the app but nothing else<br>(1) Reports of bugs, errors, or bad quality issues - i.e., something does not work in the app or is of bad quality<br>(2) Issues of the app's monetization model - i.e., complaints or issues of how the app monetizes content<br>(3) Suggestions for new features or content or revival of removed features<br>(4) Customer support issues - i.e., problems or complaints regarding customer support<br>(5) Performance issues - i.e., the app needs to much space, is to slow or similar<br>(6) Security concerns - i.e., user is concerned about their data or privacy<br>(7) Ethical concerns - i.e., user is concerned about practices in the app, fairness, discrimination<br>(8) Community related issues - user asks openly for help (no feedback at customer support) | **Section 2**<br>Expert validated data structure |
| Review text: [*RELEASE NOTE*] | **Section 3**<br>Text data for evaluation |

**Table 1 Interviews with app developers**

| Interviewee | Duration |
|---|---|
| Professional Mobile App Developer (iOS & Android), Computer Scientist (M.Sc.) | 31:13 min |
| Professional Mobile App Developer (iOS & Android), Computer Scientist (B.Sc.) | 81:26 min |
| Professional iOS Developer, Data Scientist (M.A.) | Written survey |
| Professional Mobile & Web Developer, Software Engineer (Bachelor's Degree) | 24:14 min |
| Professional Mobile App Developer (iOS & Android), Computer Scientist (B.Sc.) | 18:17 min |
| Professional Unity Game Developer and Designer (for Mobile Games), Software Engineer (Bachelor's Degree) | 26:32 min |
| Professional Unity Game Developer (for Mobile Games) | 37:49 min |

**Table 2 Model performance for identification of innovative updates**

| | | F1-Score | Precision | Recall | Accuracy | CR |
|---|---|---|---|---|---|---|
| Version First Digit (1.0 → 2.0) | *Literature* | 0.211 | 0.559 | 0.130 | 0.574 | |
| Version Second Digit (1.1 → 1.2) | | 0.472 | 0.475 | 0.470 | 0.540 | |
| Dictionary-Based | | 0.666 | 0.799 | 0.571 | 0.749 | |
| Dictionary-Character-Based | | 0.615 | 0.477 | 0.863 | 0.526 | |
| K-Nearest-Neighbor | *Classical Machine Learning Models* | 0.713 | 0.828 | 0.626 | 0.779 | |
| Logistic Regression | | 0.850 | 0.810 | 0.895 | 0.862 | |
| Naïve Bayes | | 0.835 | 0.839 | 0.831 | 0.856 | |
| Random Forest | | 0.789 | 0.740 | 0.845 | 0.802 | |
| Support Vector Machine (SVM) | | 0.849 | 0.832 | 0.868 | 0.865 | |
| Extreme Gradient Boosting (XGBoost) | | 0.834 | 0.808 | 0.863 | 0.850 | |
| Convolutional Neural Network (CNN) | *Deep Learning* | 0.828 | 0.817 | 0.838 | 0.847 | |
| BERT | *Transformer-Based PLMs* | 0.899 | 0.865 | 0.936 | 0.908 | |
| ELECTRA | | 0.889 | 0.849 | 0.934 | 0.898 | |
| RoBERTa | | 0.902 | 0.879 | 0.927 | 0.912 | |
| XLNet | | 0.887 | 0.838 | 0.943 | 0.895 | |
| Claude Haiku 3.5 | *Large Language Models* | 0.746 | 0.944 | 0.616 | 0.816 | 1.000 |
| Claude Sonnet 4 | | 0.896 | 0.938 | 0.858 | 0.913 | 0.999 |
| Mistral Small | | 0.765 | 0.934 | 0.648 | 0.826 | 0.992 |
| Mistral Large | | 0.851 | 0.930 | 0.785 | 0.880 | 0.984 |
| Mistral Small (fine-tuned) | | 0.887 | 0.888 | 0.886 | 0.901 | 0.984 |
| Mistral Large (fine-tuned) | | 0.909 | 0.891 | 0.929 | 0.919 | 0.985 |
| GPT-4.1 Nano | | 0.760 | 0.952 | 0.632 | 0.825 | 0.999 |
| GPT-4.1 Mini | | 0.869 | 0.894 | 0.845 | 0.888 | 0.987 |
| GPT-4.1 | | 0.901 | 0.923 | 0.879 | 0.915 | 0.982 |
| GPT-4.1 Nano (fine-tuned) | | 0.919 | 0.907 | 0.932 | 0.928 | 0.998 |
| GPT-4.1 Mini (fine-tuned) | | 0.914 | 0.888 | 0.941 | 0.922 | 0.992 |
| GPT-4.1 (fine-tuned) | | 0.926 | 0.894 | 0.961 | 0.933 | 0.996 |
| N | | 1,000 | 1,000 | 1,000 | 1,000 | 3,000 |

*Note.* The version-based approach follows Wen & Zhu (2018), the dictionary-based approach follows Kircher & Foerderer (2023), and the dictionary-character-based approach follows Agarwal & Kapoor (2023). With the ML-, and transformer-based classification updates of the classifications 1 or 2 were considered innovative. ML-based-classifiers were trained, respectively, transformer-based models were fine-tuned with a dataset n=2,000, with categories in distributed similarly to the validation data. Embeddings for the ML classifiers were created using TF-IDF, and embeddings for the CNN were generated using Glove 42B-300d. LLM temperature = 0.0, LLM seed = 94032. F1-score is reported for the first run. Consistency Rate (CR) is calculated based on three independent runs.

**Table 3 Model performance for update classification**

| | 1 | 2 | 3 | 4 | 5 | 6 | 7 | Macro F1 | Weighted F1 | CR |
|---|---|---|---|---|---|---|---|---|---|---|
| K-Nearest-Neighbor | 0.665 | 0.217 | 0.629 | 0.372 | 0.118 | 0.776 | 0.167 | 0.420 | 0.639 | |
| Logistic Regression | 0.793 | 0.449 | 0.673 | 0.569 | 0.211 | 0.873 | 0.000 | 0.510 | 0.759 | |
| Naïve Bayes | 0.788 | 0.496 | 0.621 | 0.563 | 0.320 | 0.878 | 0.000 | 0.524 | 0.760 | |
| Random Forest | 0.702 | 0.232 | 0.542 | 0.607 | 0.105 | 0.816 | 0.000 | 0.429 | 0.682 | |
| SVM | 0.788 | 0.463 | 0.647 | 0.583 | 0.190 | 0.873 | 0.000 | 0.506 | 0.757 | |
| XGBoost | 0.764 | 0.353 | 0.686 | 0.605 | 0.105 | 0.853 | 0.143 | 0.501 | 0.737 | |
| CNN | 0.774 | 0.349 | 0.667 | 0.457 | 0.286 | 0.854 | 0.000 | 0.484 | 0.729 | |
| BERT | 0.833 | 0.500 | 0.721 | 0.750 | 0.000 | 0.906 | 0.000 | 0.530 | 0.804 | |
| ELECTRA | 0.850 | 0.458 | 0.691 | 0.776 | 0.000 | 0.919 | 0.000 | 0.528 | 0.813 | |
| RoBERTa | 0.864 | 0.647 | 0.776 | 0.840 | 0.000 | 0.926 | 0.000 | 0.579 | 0.844 | |
| XLNet | 0.842 | 0.531 | 0.739 | 0.803 | 0.000 | 0.916 | 0.000 | 0.547 | 0.819 | |
| Claude Haiku 3.5 | 0.642 | 0.471 | 0.475 | 0.498 | 0.444 | 0.756 | 0.174 | 0.494 | 0.646 | 0.997 |
| Claude Sonnet 4 | 0.865 | 0.504 | 0.556 | 0.524 | 0.571 | 0.799 | 0.296 | 0.588 | 0.758 | 0.994 |
| Mistral Small | 0.587 | 0.441 | 0.527 | 0.466 | 0.545 | 0.738 | 0.200 | 0.501 | 0.587 | 0.971 |
| Mistral Large | 0.772 | 0.582 | 0.637 | 0.558 | 0.389 | 0.818 | 0.348 | 0.586 | 0.772 | 0.973 |
| Mistral Small (fine-tuned) | 0.821 | 0.511 | 0.752 | 0.745 | 0.500 | 0.904 | 0.276 | 0.644 | 0.812 | 0.966 |
| Mistral Large (fine-tuned) | 0.863 | 0.563 | 0.764 | 0.758 | 0.583 | 0.929 | 0.400 | 0.607 | 0.845 | 0.970 |
| GPT-4.1 Nano | 0.615 | 0.439 | 0.463 | 0.315 | 0.462 | 0.455 | 0.333 | 0.440 | 0.501 | 0.997 |
| GPT-4.1 Mini | 0.798 | 0.595 | 0.614 | 0.552 | 0.522 | 0.819 | 0.286 | 0.598 | 0.753 | 0.965 |
| GPT-4.1 | 0.849 | 0.523 | 0.741 | 0.738 | 0.435 | 0.911 | 0.400 | 0.657 | 0.825 | 0.970 |
| GPT-4.1 Nano (fine-tuned) | 0.876 | 0.681 | 0.764 | 0.797 | 0.538 | 0.940 | 0.375 | 0.710 | 0.865 | 0.993 |
| GPT-4.1 Mini (fine-tuned) | 0.882 | 0.682 | 0.779 | 0.838 | 0.522 | 0.938 | 0.400 | 0.720 | 0.870 | 0.991 |
| GPT-4.1 (fine-tuned) | 0.888 | 0.602 | 0.800 | 0.877 | 0.455 | 0.949 | 0.435 | 0.715 | 0.875 | 0.988 |
| N | 368 | 70 | 58 | 78 | 15 | 400 | 11 | 1,000 | 1,000 | 3,000 |

*Note.* ML-based-classifiers were trained, respectively, transformer-based models were fine-tuned with a dataset n=2,000, with categories in distributed similarly to the validation data. Embeddings for the ML classifiers were created using TF-IDF, and embeddings for the CNN were generated using Glove 42B-300d. LLM temperature = 0.0. F1-score is reported for the first run. Consistency Rate (CR) is based on three independent runs.

**Table 4 Model performance for review classification**

| | 0 | 1 | 2 | 3 | 4 | 5 | 6 | 7 | 8 | M F1 | W F1 | CR |
|---|---|---|---|---|---|---|---|---|---|---|---|---|
| K-Nearest-Neighbor | 0.826 | 0.431 | 0.465 | 0.092 | 0.063 | 0.000 | 0.000 | 0.000 | 0.000 | 0.208 | 0.529 | |
| Logistic Regression | 0.858 | 0.644 | 0.610 | 0.481 | 0.359 | 0.087 | 0.143 | 0.000 | 0.000 | 0.354 | 0.672 | |
| Naïve Bayes | 0.850 | 0.677 | 0.553 | 0.553 | 0.407 | 0.204 | 0.129 | 0.098 | 0.100 | 0.397 | 0.687 | |
| Random Forest | 0.813 | 0.600 | 0.602 | 0.424 | 0.511 | 0.387 | 0.000 | 0.174 | 0.100 | 0.401 | 0.645 | |
| SVM | 0.821 | 0.660 | 0.609 | 0.483 | 0.356 | 0.250 | 0.118 | 0.174 | 0.000 | 0.386 | 0.664 | |
| XGBoost | 0.846 | 0.640 | 0.573 | 0.445 | 0.356 | 0.087 | 0.000 | 0.000 | 0.000 | 0.328 | 0.655 | |
| CNN | 0.800 | 0.571 | 0.620 | 0.406 | 0.379 | 0.174 | 0.091 | 0.133 | 0.000 | 0.353 | 0.623 | |
| BERT | 0.895 | 0.733 | 0.410 | 0.551 | 0.000 | 0.000 | 0.000 | 0.000 | 0.000 | 0.288 | 0.687 | |
| ELECTRA | 0.909 | 0.767 | 0.000 | 0.611 | 0.000 | 0.000 | 0.000 | 0.000 | 0.000 | 0.254 | 0.674 | |
| RoBERTa | 0.923 | 0.758 | 0.697 | 0.700 | 0.439 | 0.000 | 0.000 | 0.100 | 0.000 | 0.402 | 0.766 | |
| XLNet | 0.917 | 0.764 | 0.651 | 0.674 | 0.642 | 0.000 | 0.000 | 0.273 | 0.000 | 0.436 | 0.766 | |
| Claude Haiku 3.5 | 0.924 | 0.829 | 0.753 | 0.741 | 0.622 | 0.609 | 0.625 | 0.429 | 0.333 | 0.652 | 0.827 | 0.989 |
| Claude Sonnet 4 | 0.914 | 0.791 | 0.780 | 0.732 | 0.621 | 0.427 | 0.632 | 0.411 | 0.444 | 0.639 | 0.813 | 1.000 |
| Mistral Small | 0.909 | 0.776 | 0.766 | 0.690 | 0.686 | 0.161 | 0.615 | 0.386 | 0.154 | 0.571 | 0.793 | 0.989 |
| Mistral Large | 0.900 | 0.779 | 0.745 | 0.724 | 0.641 | 0.415 | 0.667 | 0.346 | 0.364 | 0.620 | 0.798 | 0.911 |
| Mistral Small (fine-tuned) | 0.927 | 0.814 | 0.822 | 0.703 | 0.583 | 0.519 | 0.667 | 0.526 | 0.250 | 0.646 | 0.825 | 0.971 |
| Mistral Large (fine-tuned) | 0.937 | 0.837 | 0.814 | 0.763 | 0.476 | 0.429 | 0.643 | 0.333 | 0.500 | 0.637 | 0.836 | 0.965 |
| GPT-4.1 Nano | 0.898 | 0.764 | 0.675 | 0.524 | 0.300 | 0.261 | 0.519 | 0.160 | 0.154 | 0.473 | 0.740 | 0.995 |
| GPT-4.1 Mini | 0.905 | 0.835 | 0.739 | 0.731 | 0.651 | 0.373 | 0.629 | 0.371 | 0.256 | 0.610 | 0.812 | 0.960 |
| GPT-4.1 | 0.916 | 0.828 | 0.749 | 0.731 | 0.727 | 0.516 | 0.647 | 0.491 | 0.370 | 0.664 | 0.825 | 0.943 |
| GPT-4.1 Nano (fine-tuned) | 0.938 | 0.856 | 0.785 | 0.729 | 0.781 | 0.500 | 0.727 | 0.438 | 0.375 | 0.681 | 0.846 | 0.984 |
| GPT-4.1 Mini (fine-tuned) | 0.944 | 0.850 | 0.844 | 0.770 | 0.793 | 0.471 | 0.435 | 0.323 | 0.267 | 0.633 | 0.851 | 0.996 |
| GPT-4.1 (fine-tuned) | 0.948 | 0.872 | 0.853 | 0.769 | 0.807 | 0.500 | 0.667 | 0.467 | 0.471 | 0.706 | 0.867 | 0.997 |
| N | 475 | 240 | 92 | 146 | 31 | 19 | 13 | 19 | 12 | 1,047 | 1,047 | 3,000 |

*Note.* ML-based-classifiers were trained, respectively, transformer-based models were fine-tuned with a dataset n=2,000, with categories in distributed similarly to the validation data. Embeddings for the ML classifiers were created using TF-IDF, and embeddings for the CNN were generated using Glove 42B-300d. LLM temperature = 0.0, LLM seed = 94032. F1-score is reported for the first run. Consistency Rate (CR) is based on three independent runs.

**Table 5 Model performance for update classification with different model provider, size, age, and reasoning capabilities**

| | | (1) | | | (2) | | |
| --- | --- | --- | --- | --- | --- | --- | --- |
| | | Macro F1 | Weighted F1 | CR | Macro F1 | Weighted F1 | CR |
| Claude Opus 3 (reasoning) | Anthropic | 0.569 | 0.740 | 0.991 | 0.596 | 0.779 | 0.982 |
| Claude Opus 4 (reasoning) | | 0.511 | 0.787 | 0.852 | 0.650 | 0.823 | 0.996 |
| Claude Sonnet 3.5 | | 0.644 | 0.800 | 0.994 | 0.649 | 0.814 | 0.991 |
| Claude Sonnet 3.7 | | 0.613 | 0.771 | 0.996 | 0.638 | 0.815 | 0.987 |
| Claude Haiku 3 | | 0.481 | 0.563 | 1.000 | 0.533 | 0.732 | 1.000 |
| Ministral 3B | Mistral | 0.399 | 0.508 | 0.970 | 0.483 | 0.703 | 0.945 |
| Ministral 8B | | 0.287 | 0.378 | 0.978 | 0.402 | 0.674 | 0.983 |
| Open Mistral Nemo | | 0.277 | 0.245 | 0.880 | 0.491 | 0.695 | 0.862 |
| Mistral Medium | | 0.594 | 0.690 | 0.964 | 0.622 | 0.806 | 0.941 |
| Ministral 8B (fine-tuned) | | 0.538 | 0.756 | 0.989 | 0.571 | 0.793 | 0.994 |
| Open Mistral Nemo (fine-tuned) | | 0.485 | 0.714 | 0.945 | 0.556 | 0.788 | 0.947 |
| GPT-o3 (reasoning) | OpenAI | 0.684 | 0.809 | 0.932 | 0.693 | 0.845 | 0.935 |
| GPT-o4 Mini (reasoning) | | 0.580 | 0.734 | 0.998 | 0.693 | 0.855 | 0.978 |
| GPT-3.5 turbo | | 0.439 | 0.511 | 0.916 | 0.564 | 0.780 | 0.961 |
| GPT-4o Mini | | 0.484 | 0.571 | 0.976 | 0.576 | 0.797 | 0.934 |
| GPT-4o | | 0.602 | 0.768 | 0.956 | 0.640 | 0.813 | 0.948 |
| GPT-3.5 turbo (fine-tuned) | | 0.751 | 0.884 | 0.997 | 0.677 | 0.854 | 1.000 |
| GPT-4o Mini (fine-tuned) | | 0.744 | 0.881 | 0.993 | 0.662 | 0.858 | 0.993 |
| GPT-4o (fine-tuned) | | 0.741 | 0.889 | 0.990 | 0.678 | 0.861 | 0.993 |
| N | | 1,000 | 1,000 | 3,000 | 1,047 | 1,047 | 3,000 |

*Note.* Transformer-based models were fine-tuned with a dataset n=2,000, with categories in distributed similarly to the validation data. Claude models, Ministral 3B, Mistral Medium, GPT o3, and o4-Mini are not open for supervised fine-tuning. LLM temperature = 0.0, LLM seed = 94032.

**Table 6 Model performance for classification with different prompt designs**

| | | (1) | | | (2) | | |
|---|---|---|---|---|---|---|---|
| | | Macro F1 | Weighted F1 | CR | Macro F1 | Weighted F1 | CR |
| GPT-4.1 Nano | Few-Shot | 0.445 | 0.535 | 0.997 | 0.586 | 0.784 | 0.986 |
| GPT-4.1 Mini | | 0.621 | 0.783 | 0.954 | 0.610 | 0.813 | 0.927 |
| GPT-4.1 | | 0.658 | 0.822 | 0.975 | 0.662 | 0.831 | 0.952 |
| GPT-4.1 Nano (fine-tuned) | | 0.650 | 0.856 | 0.995 | 0.660 | 0.842 | 0.987 |
| GPT-4.1 Mini (fine-tuned) | | 0.724 | 0.866 | 0.999 | 0.663 | 0.856 | 0.988 |
| GPT-4.1 (fine-tuned) | | 0.739 | 0.881 | 0.987 | 0.695 | 0.866 | 0.985 |
| GPT-4.1 Nano | Chain-Of-Thought (Automatic) | 0.428 | 0.485 | 0.996 | 0.482 | 0.753 | 0.991 |
| GPT-4.1 Mini | | 0.596 | 0.751 | 0.962 | 0.621 | 0.806 | 0.946 |
| GPT-4.1 | | 0.647 | 0.821 | 0.977 | 0.672 | 0.837 | 0.961 |
| GPT-4.1 Nano (fine-tuned) | | 0.593 | 0.858 | 0.993 | 0.653 | 0.840 | 0.984 |
| GPT-4.1 Mini (fine-tuned) | | 0.721 | 0.869 | 0.999 | 0.645 | 0.843 | 0.992 |
| GPT-4.1 (fine-tuned) | | 0.710 | 0.874 | 0.991 | 0.702 | 0.860 | 0.991 |
| GPT-4.1 Nano | Chain-Of-Thought (Manual) | 0.411 | 0.415 | 0.997 | 0.470 | 0.722 | 0.980 |
| GPT-4.1 Mini | | 0.577 | 0.723 | 0.957 | 0.621 | 0.798 | 0.958 |
| GPT-4.1 | | 0.631 | 0.808 | 0.976 | 0.674 | 0.831 | 0.926 |
| GPT-4.1 Nano (fine-tuned) | | 0.554 | 0.841 | 0.994 | 0.662 | 0.834 | 0.982 |
| GPT-4.1 Mini (fine-tuned) | | 0.719 | 0.872 | 0.995 | 0.647 | 0.850 | 0.993 |
| GPT-4.1 (fine-tuned) | | 0.722 | 0.876 | 0.983 | 0.691 | 0.863 | 0.992 |
| GPT-4.1 Nano | Chain-Of-Thought (Contrastive) | 0.341 | 0.355 | 0.992 | 0.468 | 0.725 | 0.978 |
| GPT-4.1 Mini | | 0.603 | 0.737 | 0.947 | 0.616 | 0.802 | 0.956 |
| GPT-4.1 | | 0.671 | 0.823 | 0.985 | 0.684 | 0.831 | 0.940 |
| GPT-4.1 Nano (fine-tuned) | | 0.649 | 0.840 | 0.993 | 0.676 | 0.841 | 0.982 |
| GPT-4.1 Mini (fine-tuned) | | 0.729 | 0.873 | 0.996 | 0.645 | 0.854 | 0.994 |
| GPT-4.1 (fine-tuned) | | 0.748 | 0.882 | 0.987 | 0.710 | 0.867 | 0.990 |
| N | | 1,000 | 1,000 | 3,000 | 1,047 | 1,047 | 3,000 |

*Note.* Transformer-based models were fine-tuned with a dataset n=2,000, with categories in distributed similarly to the validation data. LLM temperature = 0.0, LLM seed = 94032. See Online Appendix C1 for prompts.

**Figure 6 Model performance for update classification with different temperatures**

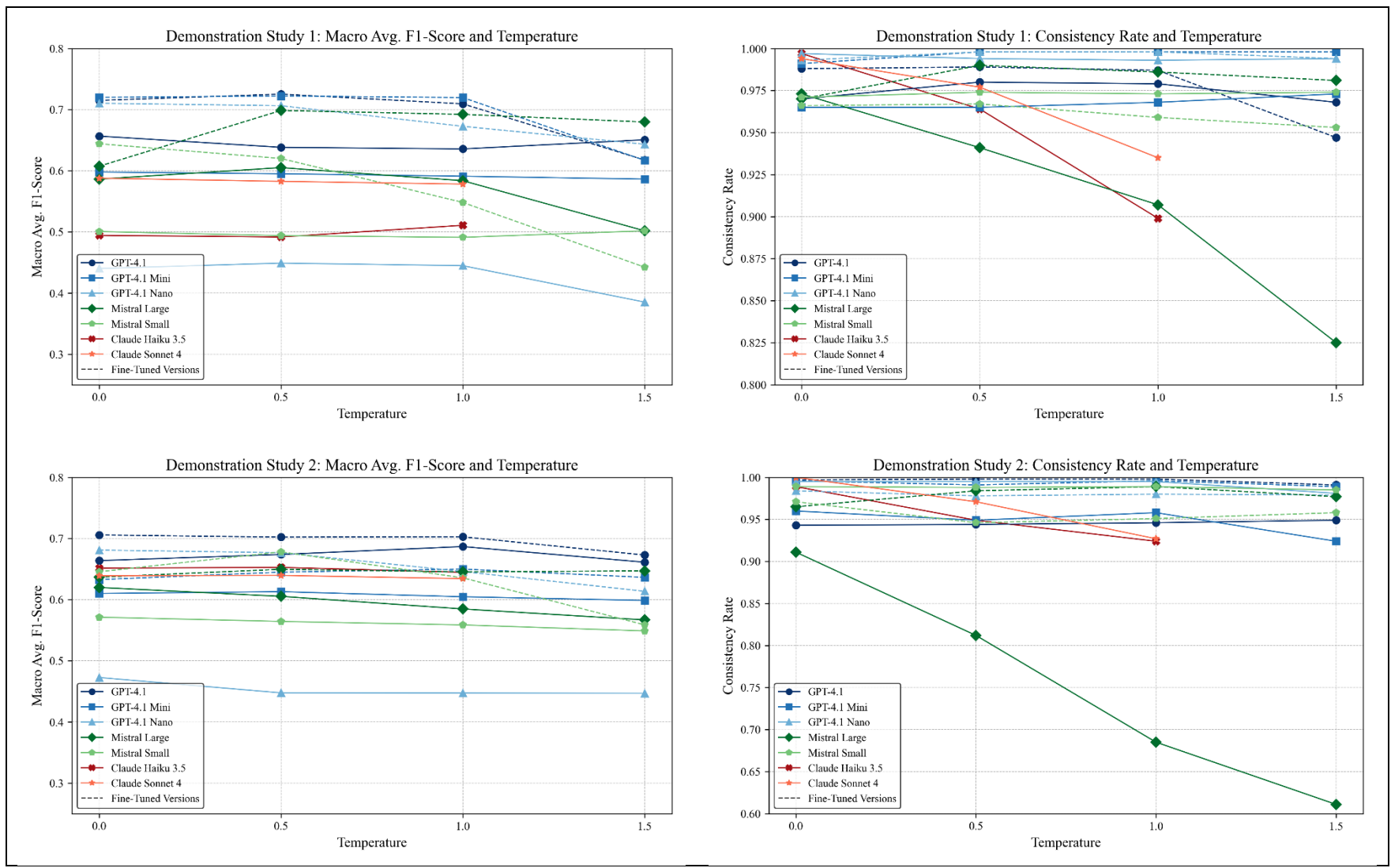

*Note.* LLMs were fine-tuned with a dataset n=2,000, with categories in distributed similarly to the validation data. F1-score is reported for the first run. Consistency Rate (CR) is based on three independent runs. Y-axes are scaled to fit best and worst model values. The maximum temperature for Claude models is limited to 1.0 per provider documentation.

**Figure 7 Model performance for update classification with different training data distribution and size**

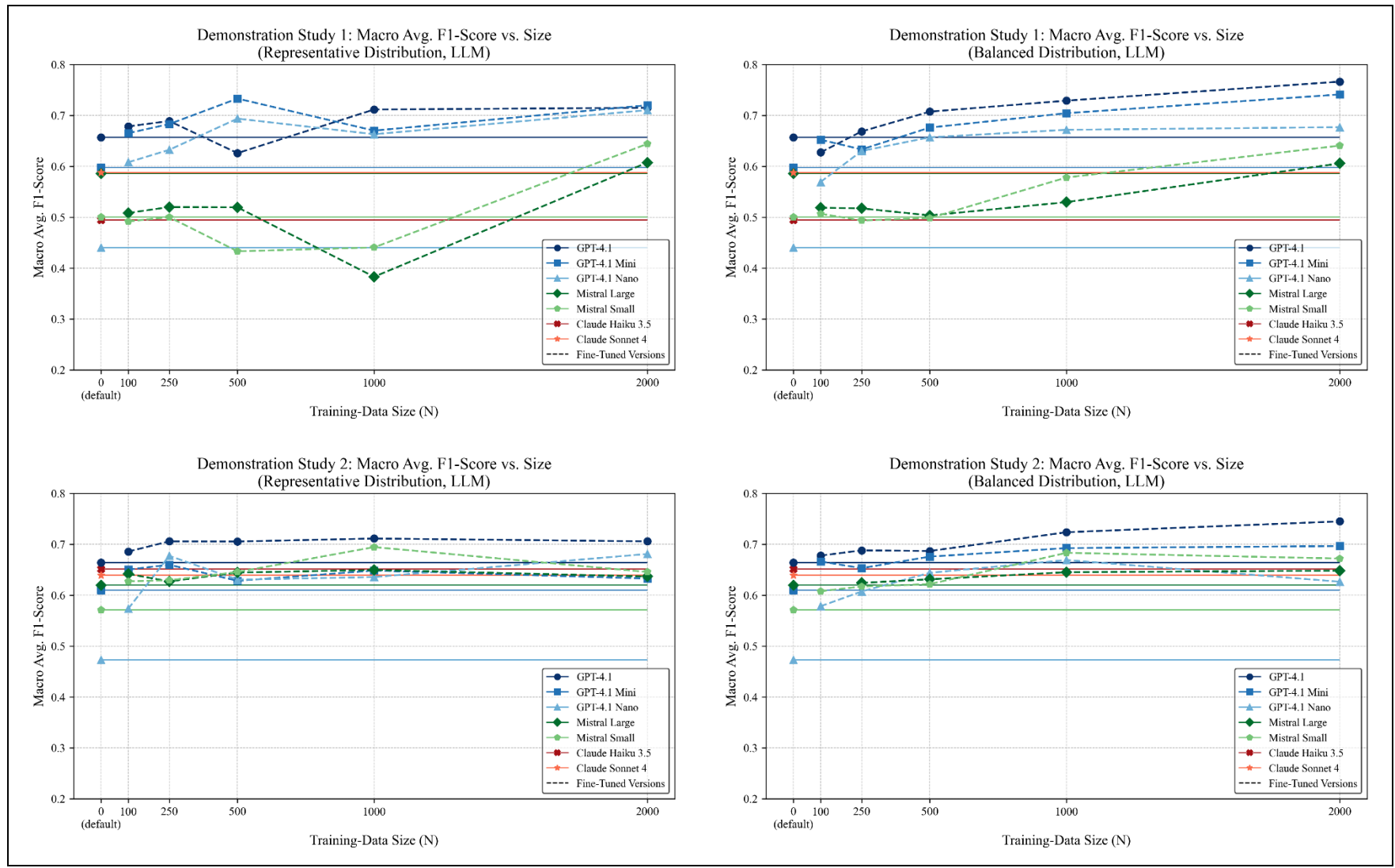

*Note.* F1-score is reported for the first run. Y-axes are scaled to fit best and worst model values.

**Online Appendix**

AI-Based Measurement of Innovation: Mapping Expert Insight into Large Language Model Applications

**Paper Companion Repository:** https://github.com/robi979/AI-Innovation-Measurement

**Appendix A. Hyperparameter Tuning**

**A1  Tuning Numerical Text Representations and Hyperparameters for Classical Machine Learning Classifiers**

To identify the optimal model configuration for update classification, we systematically tuned both feature representations and classifier hyperparameters using an automated and reproducible pipeline via the *sckit-learn* (Pedregosa et al. 2011) and *xgboost* (Chen and Guestrin 2016) libraries. For the main analysis in Section 4 of the paper, we used representative training sets consisting of n = 2,000 release notes and labels for Demonstration Study 1, and n = 2,000 reviews (including title and body) and labels for Demonstration Study 2. For the additional analysis in Section 5.4 of the paper, we varied the training data distribution (balanced) and size (n = 100, 250, 500, 1,000) in both studies. To ensure compatibility with *scikit-learn*, labels in Study 1 were zero-indexed. For the multi-label classification task in Study 2, labels were transformed into multi-hot vectors. A fixed hold-out validation set (n = 1,000) was reserved for final evaluation.

Two feature representation approaches were evaluated. First, *TF-IDF* vectorization was applied, with the following hyperparameters included in the search: *max_features* (random integers between 2,000 and 10,000), *ngram_range* ((1,1), (1,2), or (1,3)), *min_df* (1-4), and *max_df* (uniformly between 0.6 and 1.0). Second, distributed representations were obtained by averaging pretrained *GloVe* word embeddings, considering four variants: *6B-100d*, *6B-300d*, *42B-300d*, and *840B-300d* (Pennington et al. 2014). Document vectors were precomputed and cached for each embedding.

For each representation, we evaluated *Random Forest*, *Logistic Regression*, *K-Nearest Neighbors*, *Support Vector Machine*, and *XGBoost* classifiers. *Multinomial Naive Bayes* was included for *TF-IDF* features but excluded for *GloVe* representations as the algorithm is designed specifically for discrete feature counts (such as those generated by *TF-IDF*) and does not perform reliably on continuous or dense embeddings like *GloVe*. The hyperparameter ranges were as follows:

- Random Forest: *n_estimators* (50-500), *max_depth* (5-100), *min_samples_split* (2-20), *min_samples_leaf* (1-10), *max_features* (0.1-1.0), *criterion* ('gini', 'entropy'), *bootstrap* (True, False)

- Logistic Regression: *C* (0.01-10), *penalty* ('l2'), *solver* ('lbfgs', 'saga')

- K-Nearest Neighbors: *n_neighbors* (1-30), *weights* ('uniform', 'distance'), *leaf_size* (10-50)

- Multinomial Naive Bayes: *alpha* (1e-6-1.0), *fit_prior* (True, False)

- Support Vector Machine: *C* (0.1-10), *kernel* ('linear', 'rbf', 'poly'), *gamma* ('scale', 'auto'), *degree* (2-5)
- XGBoost: *n_estimators* (50-500), *max_depth* (3-10), *learning_rate* (0.01-0.3), *subsample* (0.5-1.0), *colsample_bytree* (0.5-1.0)

Hyperparameter optimization was conducted using scikit-learn's *RandomizedSearchCV* with 50 randomly sampled parameter sets per classifier. For *TF-IDF* pipelines, both vectorizer and classifier parameters were tuned jointly. For *GloVe*, only classifier parameters were tuned using the precomputed document vectors. All searches used 3-fold stratified cross-validation and optimized the macro-averaged F1 score, with a fixed random seed (random_state=94032) for full reproducibility. Computation was parallelized for efficiency.

For Demonstration Study 1 (single-label), standard scikit-learn classifiers were used. For Demonstration Study 2 (multi-label), all classifiers were wrapped with OneVsRestClassifier, and multi-hot encoded targets were used as labels. The best configuration for each representation, embedding, classifier, and training set was selected based on cross-validation performance, retrained on the full training data, and evaluated on the hold-out validation set.

----------------------------------------

Insert Table A1 about here

----------------------------------------

Table A1 summarizes the best-performing hyperparameter configurations identified via randomized search for the machine learning classifiers used in the main analyses in Section 4 of the paper. For each classifier, we report the macro-averaged F1 scores from both cross-validated training and evaluation on the hold-out validation set. Comprehensive details including all tested configurations, cross-validation results, search parameters, predictions on the validation set, and summary metrics are provided in the online repository to ensure full reproducibility.

## A2  Tuning Hyperparameters for Convolutional Neural Network Classifier

To identify optimal configurations for neural network based text classification, we performed automated hyperparameter tuning of a convolutional neural network for sentence classification (*TextCNN*) with pretrained embeddings (Kim 2014). Data preparation was not repeated, the same training and validation splits as for the machine learning classifiers were used for the *TextCNN* runs.

For each dataset, the pipeline consisted of a custom tokenizer and the CNN implemented via the *skorch scikit-learn* library (Tietz et al. 2017). Tokens were mapped to 300-dimensional *GloVe* embeddings (42B-300d), which were preloaded and assigned to each vocabulary item. The following parameter ranges were included in the search:

- Tokenizer: *num_words* (10,000-30,000), *seq_len* (100-300)
- CNN architecture: *n_filters* (64-256), *kernel_size* (3-8), *dense_units* (32-256), *dropout* (0.1-0.5), *trainable* (True, False)
- Training: *batch_size* (32-128), *max_epochs* (5-15)

The hyperparameters were jointly optimized using scikit-learn's *RandomizedSearchCV*, sampling 30 parameter configurations per run.[1] All searches used 3-fold stratified cross-validation and optimized the macro-averaged F1 score, with a fixed random seed (random_state=94032) for full reproducibility. Computation was parallelized for efficiency.

In Demonstration Study 1, the network was trained using categorical cross-entropy loss for single-label classification. In Demonstration Study 2, a multi-label extension of CNN with a binary cross-entropy loss function (*BCEWithLogitsLoss*) was employed to accommodate multi-hot target vectors. The configurations achieving the best mean cross-validated macro-F1 score was retrained on the full training data and evaluated on the hold-out validation set.

----------------------------------------

Insert Table A2 about here

----------------------------------------

Table A2 summarizes the best-performing hyperparameter configuration identified via *RandomizedSearchCV* for the Convolutional Neural Network used in the main analyses in Section 4 of the paper. Macro-averaged F1 scores from both cross-validated training and evaluation on the hold-out validation set are reported. Comprehensive details including all tested configurations, cross-validation results, search parameters, predictions on the validation set, and summary metrics are provided in the online repository to ensure full reproducibility.

---

[1] Given the increased computational effort associated with neural model training, 30 randomized hyperparameter configurations were evaluated for each CNN pipeline, compared to 50 for classical models.

**A3 Tuning Hyperparameters for Pre-Trained Language Models**

To identify optimal configurations for transformer-based text classification, we fine-tuned several pretrained language models (PLMs) on the same training and validation splits as the classical machine learning and CNN experiments.

We evaluated four widely used pretrained transformer models *BERT-base-cased* (Devlin et al. 2019), *RoBERTa-base* (Liu et al. 2019), *XLNet-base-cased* (Yang et al. 2019), and *ELECTRA-base-discriminator* (Clark et al. 2020) —using the Hugging Face *transformers* library (Wolf et al. 2020). For each model, the associated tokenizer was used to preprocess text, truncating or padding each sample to a fixed *maximum_length* (=512 tokens). In Demonstration Study 1, labels were zero-indexed, and models were trained using a standard cross-entropy objective. In Demonstration Study 2, labels were transformed to multi-hot vectors, and models were configured for multi-label classification using a sigmoid output and binary cross-entropy loss.

Hyperparameter optimization was performed using *Optuna* (Akiba et al. 2019) with 5 trials per model and training set. For each trial, the following parameters were tuned: learning_rate (log-uniform between 1e-5 and 5e-5), batch_size (8 or 16), num_train_epochs (3–5), and weight_decay (0.0–0.01). Each configuration was trained and evaluated using the Hugging Face *Trainer* framework, and the macro-averaged F1 score on the validation set was used as the selection criterion. All experiments were seeded (*random_state*=94032) for reproducibility. For each model and training split, the configuration with the highest validation macro-F1 score was selected, and saved for downstream evaluation.

----------------------------------------

Insert Table A3 about here

----------------------------------------

Table A3 summarizes the best-performing hyperparameter configurations identified via *Optuna* trials for the PLMs used in the main analyses in Section 4 of the paper. For each model, we report the macro-averaged F1 scores from both cross-validated training and evaluation on the hold-out validation set. Comprehensive details including all tested configurations, cross-validation results, search parameters, predictions on the validation set, and summary metrics are provided in the online repository to ensure full reproducibility.

**Appendix B. Large Language Model Inference via Batch API**

**B1  Batch Processing**

To evaluate large language model (LLM) performance at scale in both Demonstration Studies, we leveraged batch inference APIs from all three model providers included in the experiments—OpenAI, Mistral, and Anthropic—across all design decision variations.

For each provider, a systematic procedure was implemented to generate, upload, and execute batch requests for both base and fine-tuned models, adhering to the respective provider's API documentation.[2] The set of evaluated models included:

- OpenAI: *GPT-3.5-turbo-0125* (default & fine-tuned), *GPT-4o-mini-2024-07-18* (default & fine-tuned), *GPT-4o-2024-08-06* (default & fine-tuned), *GPT-4.1-2025-04-14* (default & fine-tuned), *GPT-4.1-mini-2025-04-14* (default & fine-tuned), *GPT-4.1-nano-2025-04-14* (default & fine-tuned), *o3-2025-04-16* (default), *o4-mini-2025-04-16* (default)

- Mistral: *mistral-large-2411* (default & fine-tuned), *mistral-medium-2505* (default), *mistral-small-2503* (default & fine-tuned), *open-mistral-nemo-2407* (default & fine-tuned), *ministral-8b-2410* (default & fine-tuned), *ministral-3b-2410* (default)

- Anthropic: *claude-sonnet-4-20250514* (default), *claude-3-7-sonnet-20250219* (default), *claude-3-5-haiku-20241022* (default), *claude-3-5-sonnet-20241022* (default), *claude-3-haiku-20240307* (default), *claude-opus-4-20250514* (default), *claude-3-opus-20240229* (default)

Fine-tuned models were trained on datasets (distribution and size) similar to those in sections 1-3 above but included the resepective prompts. Further design decisions reported in section 5 of the paper included *prompt* engineering and *temperature* setting. As such we itereated each model over the *prompt* templates (default, few-shot, automatic-chain-of-thought, manual-chain-of-thought, contrastive-chain-of-thought) and *temperature* settings (0, 0.5, 1.0, 1.5). For Anthropic models, the maximum *temperature* was limited to 1.0 per provider documentation. For each model-prompt-temperature combination, three repeated runs were performed on the full validation dataset, using a fixed *seed* (OpenAI and Anthropic) or *random_*seed (Mistral) parameter set to 94032 for reproducibility. For all batch jobs, *max_completion_tokens* (OpenAI) or *max_tokens* (Mistral/Anthropic) was set to 1000.

---

[2] See https://platform.openai.com/docs/guides/batch (Retrieved July 11, 2025),
https://docs.mistral.ai/capabilities/batch/ (Retrieved July 11, 2025), and
https://docs.anthropic.com/en/docs/build-with-claude/batch-processing (Retrieved July 11, 2025) for details.

For OpenAI's reasoning models (i.e., *o3-2025-04-16* and *o4-mini-2025-04-16*), the *temperature* parameter was omitted in all batch requests, as these models do not accept temperature control. Additionally, for these models, the *reasoning_effort* parameter was set to 'low'.

After all batch files were constructed, they were uploaded using the respective provider's API. For OpenAI and Mistral, each uploaded batch file produced a file identifier, which was then referenced when creating a batch classification job, specifying the model and endpoint (*/v1/chat/completions*), along with metadata for job tracking. For Anthropic, batch submission was performed via the *messages.batches.create* endpoint, directly uploading a list of parameterized requests. Unique identifiers (*custom_id*) were assigned to every request for traceability and later result matching.

## B2 Post-processing and Output Cleaning

To evaluate large language model (LLM) performance at scale in both Demonstration Studies, we leveraged batch inference APIs from all three model providers included in the experiments—OpenAI, Mistral, and Anthropic—across all design decision configurations.

To ensure comparability and accurate evaluation of LLM predictions, a standardized post-processing pipeline was applied to all raw outputs from the batch API runs across providers. For each provider (OpenAI, Mistral, Anthropic), provider-specific parsing functions were implemented to extract the predicted content and align it with the corresponding sample identifiers. Within each output file, the *custom_id* field was parsed to reconstruct the relevant design configuration including the model name, prompt template, temperature setting, run index, and sample identifier. Further the raw batch outputs (*message*) were parsed to extract the predicted content. The extracted predictions were mapped back to the validation set.

Given the potential variability in free-form LLM outputs, especially under more complex prompts or increased temperature values, we implemented an output cleaning procedure to reliably extract class labels. In particular, we used *regular expressions* to accommodate a wide range of output formats, correctly extracting single integer predictions, multiple class outputs separated by semicolons (as in Demonstration Study 2), as well as labels embedded in extra texts and further non-numeric characters such as parentheses or stray punctuation. Any invalid, missing, or out-of-range predictions (i.e., anything not mapping to the defined target labels of the Demonstration Study) were assigned a fallback class label to ensure these were penalized in downstream performance evaluation.

**Appendix C. Design Decisions in Demonstration Studies**

**C1 Prompt Engineering Techniques (Demonstration Study 1)**

*Few-Shot Prompt*

ONLY provide the category number (1-7) in response. Determine the category for the following app update text.
If multiple categories seem applicable, always choose the lowest category number (1<2<3<4<5<6<7):

   (1) Novel features - Introducing or enhancing significant functionalities that modify user experience
   (2) Content extensions – Adding or expanding curated or user-facing content
   (3) Platform and device support – Enabling compatibility with new OS versions, development kits, device types, or hardware features
   (4) Specific fixes of bugs - Addressing distinct known issues
   (5) Privacy & Security - Strengthening user data protection, permissions, or security protocols
   (6) General Improvements – Tweaks addressing undisclosed bug fixes, performance improvements, minor changes, or UI/UX polish
   (7) Marketing & Branding – Changes purely about promotional or visual branding assets

EXAMPLES:
Updates: 'new style', 'new features', 'new interface'
Response: 1

Updates: 'optimized performance', 'new levels', 'new stage added'
Response: 2

Updates: 'updated for iOS 9', 'resolves iOS 9 issues', 'adapt to iOS 12 system'
Response: 3

Updates: 'fix for push-messaging', 'fixed game freeze bug', 'fixed a crash that could occur when starting the game'
Response: 4

Updates: 'security updates', 'terms of use updated', 'added prompt to ask permission for App Tracking Transparency'
Response: 5

Updates: 'bug fixes', 'stability improvements', 'improved feeds'
Response: 6

Updates: 'new app icon introduced', 'screenshots for App Store added', 'Spring has sprung and it's time to get floral! Collect as many blossoms as you can; the more you collect, the bigger the reward!'
Response: 7

*Chain-of-Thought (Automatic) Prompt*

ONLY provide the category number (1-7) in response. Determine the category for the following app update text.
If multiple categories seem applicable, always choose the lowest category number (1<2<3<4<5<6<7):

   (1) Novel features - Introducing or enhancing significant functionalities that modify user experience

(2) Content extensions – Adding or expanding curated or user-facing content
(3) Platform and device support – Enabling compatibility with new OS versions, development kits, device types, or hardware features
(4) Specific fixes of bugs - Addressing distinct known issues
(5) Privacy & Security - Strengthening user data protection, permissions, or security protocols
(6) General Improvements – Tweaks addressing undisclosed bug fixes, performance improvements, minor changes, or UI/UX polish
(7) Marketing & Branding – Changes purely about promotional or visual branding assets

Let's think step by step.

*Chain-of-Thought (Manual) Prompt*

ONLY provide the category number (1-7) in response. Determine the category for the following app update text.
If multiple categories seem applicable, always choose the lowest category number (1<2<3<4<5<6<7):

(1) Novel features - Introducing or enhancing significant functionalities that modify user experience
(2) Content extensions – Adding or expanding curated or user-facing content
(3) Platform and device support – Enabling compatibility with new OS versions, development kits, device types, or hardware features
(4) Specific fixes of bugs - Addressing distinct known issues
(5) Privacy & Security - Strengthening user data protection, permissions, or security protocols
(6) General Improvements – Tweaks addressing undisclosed bug fixes, performance improvements, minor changes, or UI/UX polish
(7) Marketing & Branding – Changes purely about promotional or visual branding assets

Let's think step by step.

ANALYZE STEP BY STEP:

1. Read the app update text carefully
2. Identify the main purpose of the update
3. Check the different categories starting from 1 to 6 and determine which category applies best
4. If multiple categories apply, choose the category with the lowest number
5. Provide the number of the category as response

EXAMPLE:

Update text: 'game runs faster and more levels have been added'

1. Reading of update text 'game runs faster and more levels have been added'
2. Main purpose of 'game runs faster' is performance improvement and purpose of 'more levels have been added' is extensions of app content
3. Performance improvement aligns with category 6, extensions of app content aligns with category 2
4. Lower category number 2 is chosen
5. 2 is provided as response"

*Chain-of-Thought (Contrastive) Prompt*

ONLY provide the category number (1-7) in response. Determine the category for the following app update text.

If multiple categories seem applicable, always choose the lowest category number (1<2<3<4<5<6<7):

(1) Novel features - Introducing or enhancing significant functionalities that modify user experience
(2) Content extensions – Adding or expanding curated or user-facing content
(3) Platform and device support – Enabling compatibility with new OS versions, development kits, device types, or hardware features
(4) Specific fixes of bugs - Addressing distinct known issues
(5) Privacy & Security - Strengthening user data protection, permissions, or security protocols
(6) General Improvements – Tweaks addressing undisclosed bug fixes, performance improvements, minor changes, or UI/UX polish
(7) Marketing & Branding – Changes purely about promotional or visual branding assets

Let's think step by step.

ANALYZE STEP BY STEP:

1. Read the app update text carefully
2. Identify the main purpose of the update
3. Check the different categories starting from 1 to 6 and determine which category applies best
4. If multiple categories apply, choose the category with the lowest number
5. Provide the number of the category as response

CORRECT EXAMPLE:

Update text: 'game runs faster and more levels have been added'

1. Reading of update text 'game runs faster and more levels have been added'
2. Main purpose of 'game runs faster' is performance improvement and purpose of 'more levels have been added' is extensions of app content
3. Performance improvement aligns with category 6, extensions of app content aligns with category 2
4. Lower category number 2 is chosen
5. 2 is provided as response"

WRONG EXAMPLE:

Update text: 'game runs faster and more levels have been added'

1. Reading of update text 'game runs faster and more levels have been added'
2. Main purpose of 'game runs faster' is extensions/additions of app content and purpose of 'more levels have been added' is privacy & security
3. Extensions/additions of app content aligns with category 6, privacy and security aligns with category 2
4. Lower category number 2 is chosen
5. 1 is provided as response

## C2 Prompt Engineering Techniques (Demonstration Study 2)

*Few-Shot Prompt*

"ONLY provide a number (0-8) in response. Categorize the following app review text by assigning the most fitting category/categories out of the following nine categories. If the text contains elements from multiple categories, provide the categories separated by ; .

- (0) User Opinion without specific reports, issues, suggestions - i.e., review only about good/bad perception of the app but nothing else
- (1) Reports of bugs, errors, or bad quality issues - i.e., something does not work in the app or is of bad quality
- (2) Issues of the app's monetization model - i.e., complaints or issues of how the app monetizes content
- (3) Suggestions for new features or content or revival of removed features
- (4) Customer support issues - i.e., problems or complaints regarding customer support
- (5) Performance issues - i.e., the app needs too much space, is too slow or similar
- (6) Security concerns - i.e., user is concerned about their data or privacy
- (7) Ethical concerns - i.e., user is concerned about practices in the app, fairness, discrimination
- (8) Community related issues - user asks openly for help (no feedback at customer support)

EXAMPLES:
Review texts: 'I love this game', 'clear, friendly, effective', 'horrible app; if I could give these creators 0 stars, I would'
Response: 0

Review texts: 'I'm getting error messages when trying to use the app', 'makes my phone incredibly hot, then the game crashes', 'I am unable to access the app'
Response: 1

Review texts: 'the items should be at a lower price', 'the apps are unbelievably disturbing', 'the app hides the ability to cancel the subscription'
Response: 2

Review texts: 'add home screen widgets', 'please support on iPad', 'create stories'
Response: 3

Review texts: 'there is no solution even though I reported the issue already', 'nobody replies from support', 'I cannot contact the support'
Response: 4

Review texts: 'the app drains 50% battery overnight', 'updating the app takes too long', 'the game lagged a bit'
Response: 5

Review texts: 'do not trust anything about this app, it's a scam', 'they are selling our privacy', 'app reads and saves photos'
Response: 6

Review texts: 'the community tends to be very toxic towards each other', 'the app is debased by its selective censorship', 'this app clearly promotes sex trafficking'
Response: 7

Review texts: 'how can I change the language', 'hey guys, I have no idea what the color paints are for', 'please unfreeze my account and support me'
Response: 8

*Chain-of-Thought (Automatic) Prompt*

"Only provide a number (0-8) as response. Categorize the following app review text based on the most applicable out of the following 9 categories. If multiple categories apply, provide the categories separated by ; .

- (0) User Opinion without specific reports, issues, suggestions - i.e., review only about good/bad perception of the app but nothing else
- (1) Reports of bugs, errors, or bad quality issues - i.e., something does not work in the app or is of bad quality
- (2) Issues of the app's monetization model - i.e., complaints or issues of how the app monetizes content
- (3) Suggestions for new features or content or revival of removed features
- (4) Customer support issues - i.e., problems or complaints regarding customer support
- (5) Performance issues - i.e., the app needs too much space, is too slow or similar
- (6) Security concerns - i.e., user is concerned about their data or privacy
- (7) Ethical concerns - i.e., user is concerned about practices in the app, fairness, discrimination
- (8) Community related issues - user asks openly for help (no feedback at customer support)

Let's think step by step.

*Chain-of-Thought (Manual) Prompt*

"Only provide a number (0-8) as response. Categorize the following app review text based on the most applicable out of the following 9 categories. If multiple categories apply, provide the categories separated by ; .

- (0) User Opinion without specific reports, issues, suggestions - i.e., review only about good/bad perception of the app but nothing else
- (1) Reports of bugs, errors, or bad quality issues - i.e., something does not work in the app or is of bad quality
- (2) Issues of the app's monetization model - i.e., complaints or issues of how the app monetizes content
- (3) Suggestions for new features or content or revival of removed features
- (4) Customer support issues - i.e., problems or complaints regarding customer support
- (5) Performance issues - i.e., the app needs too much space, is too slow or similar
- (6) Security concerns - i.e., user is concerned about their data or privacy
- (7) Ethical concerns - i.e., user is concerned about practices in the app, fairness, discrimination
- (8) Community related issues - user asks openly for help (no feedback at customer support)

ANALYZE STEP BY STEP:

1. Read the app review text carefully
2. Identify the main purpose of the review
3. Check the different categories starting from 0 to 8 and determine which category/categories applies/apply best
4. Provide the number(s) of the category/categories as response

EXAMPLE:

Review text: 'Trying to get an answer about anonymity. Can a winner stay anonymous? They have my name and address. Emailed for an answer- no response. Reading of update text 'game runs faster and more levels have been added'

1.     Reading of update text 'Trying to get an answer about anonymity. Can a winner stay anonymous? They have my name and address. Emailed for an answer- no response.'
2.     Main purpose of 'trying to get an answer' and 'Emailed for an answer - no response' is customer support issue and purpose of 'anonymity' and 'They have my name and address' is security concerns
3.     Customer support issue aligns with category 4, security concerns aligns with category 6
4.     4;6 are provided as response

*Chain-of-Thought (Contrastive) Prompt*

"Only provide a number (0-8) as response. Categorize the following app review text based on the most applicable out of the following 9 categories. If multiple categories apply, provide the categories separated by ; .

(0) User Opinion without specific reports, issues, suggestions - i.e., review only about good/bad perception of the app but nothing else
(1) Reports of bugs, errors, or bad quality issues - i.e., something does not work in the app or is of bad quality
(2) Issues of the app's monetization model - i.e., complaints or issues of how the app monetizes content
(3) Suggestions for new features or content or revival of removed features
(4) Customer support issues - i.e., problems or complaints regarding customer support
(5) Performance issues - i.e., the app needs too much space, is too slow or similar
(6) Security concerns - i.e., user is concerned about their data or privacy
(7) Ethical concerns - i.e., user is concerned about practices in the app, fairness, discrimination
(8) Community related issues - user asks openly for help (no feedback at customer support)

ANALYZE STEP BY STEP:

1. Read the app review text carefully
2. Identify the main purpose of the review
3. Check the different categories starting from 0 to 8 and determine which category/categories applies/apply best
4. Provide the number(s) of the category/categories as response

CORRECT EXAMPLE:

Review text: 'Trying to get an answer about anonymity. Can a winner stay anonymous? They have my name and address. Emailed for an answer- no response. Reading of update text 'game runs faster and more levels have been added'

1.     Reading of update text 'Trying to get an answer about anonymity. Can a winner stay anonymous? They have my name and address. Emailed for an answer- no response.'
2.     Main purpose of 'trying to get an answer' and 'Emailed for an answer - no response' is customer support issue and purpose of 'anonymity' and 'They have my name and address' is security concerns
3.     Customer support issue aligns with category 4, security concerns aligns with category 6
4.     4;6 are provided as response

WRONG EXAMPLE:

Review text: 'Trying to get an answer about anonymity. Can a winner stay anonymous? They have my name and address. Emailed for an answer- no response. Reading of update text 'game runs faster and more levels have been added'

1. Reading of update text 'Trying to get an answer about anonymity. Can a winner stay anonymous? They have my name and address. Emailed for an answer- no response.'
2. Main purpose of 'trying to get an answer' and 'Emailed for an answer - no response' is security concerns and purpose of 'anonymity' and 'They have my name and address' is ethical concerns
3. Security concerns aligns with category 4, ethical concerns aligns with category 6
4. 1;7 are provided as response

## C3 Additional Analyses on the Role of Training Data Distribution and Data Size

-----------------------------------------

Insert Figure C1 about here

-----------------------------------------

**Table A1 Hyperparameter configurations and performance evaluation for ML classifiers for demonstration studies 1 & 2**

| | | (1) Best Performing Hyperparameter Configuration | (1) Best CV Macro F1 | (1) HO Macro F1 | (2) Best Performing Hyperparameter Configuration | (2) Best CV Macro F1 | (2) HO Macro F1 |
|---|---|---|---|---|---|---|---|
| K-Nearest-Neighbor | TF-IDF | {"leaf_size": 38, "n_neighbors": 8, "weights": "distance", "max_df": 0.9240475867068707, "max_features": 8335, "min_df": 1, "ngram_range": [1, 3]} | 0.486 | 0.420 | {"leaf_size": 37, "n_neighbors": 18, "weights": "uniform", "max_df": 0.9882818784859139, "max_features": 7518, "min_df": 1, "ngram_range": [1, 1]} | 0.198 | 0.208 |
| Logistic Regression | | {"C": 6.828621268586129, "penalty": "l2", "solver": "lbfgs", "max_df": 0.996760773680571, "max_features": 9801, "min_df": 3, "ngram_range": [1, 1]} | 0.543 | 0.510 | {"C": 8.839465761662865, "penalty": "l2", "solver": "saga", "max_df": 0.9958605089165147, "max_features": 4093, "min_df": 4, "ngram_range": [1, 1]} | 0.302 | 0.354 |
| Naïve Bayes | | {"alpha": 0.11917071876314952, "fit_prior": false, "max_df": 0.8326423978632586, "max_features": 9392, "min_df": 3, "ngram_range": [1, 2]} | 0.543 | 0.524 | {"alpha": 0.11917071876314952, "fit_prior": false, "max_df": 0.8326423978632586, "max_features": 9392, "min_df": 3, "ngram_range": [1, 2]} | 0.380 | 0.397 |
| Random Forest | | {"bootstrap": false, "criterion": "entropy", "max_depth": 96, "max_features": 0.6779687087336601, "min_samples_leaf": 3, "min_samples_split": 11, "n_estimators": 448, "max_df": 0.8260307570294546, "max_features": 2630, "min_df": 1, "ngram_range": [1, 2]} | 0.477 | 0.429 | {"bootstrap": false, "criterion": "gini", "max_depth": 95, "max_features": 0.2077430655332096, "min_samples_leaf": 1, "min_samples_split": 8, "n_estimators": 198, "max_df": 0.6627743230167918, "max_features": 5848, "min_df": 1, "ngram_range": [1, 3]} | 0.344 | 0.401 |
| SVM | | {"C": 8.73445728344814, "degree": 2, "gamma": "auto", "kernel": "linear", "max_df": 0.685093473488098, "max_features": 5666, "min_df": 1, "ngram_range": [1, 2]} | 0.590 | 0.506 | {"C": 6.918621268586129, "degree": 2, "gamma": "scale", "kernel": "linear", "max_df": 0.6041717425276485, "max_features": 5270, "min_df": 4, "ngram_range": [1, 1]} | 0.393 | 0.386 |
| XGBoost | | {"colsample_bytree": 0.6965180821364413, "learning_rate": 0.07583861356997272, "max_depth": 3, "n_estimators": 445, "subsample": 0.6734468864139178, "max_df": 0.6783234961893887, "max_features": 8100, "min_df": 3, "ngram_range": [1, 3]} | 0.512 | 0.501 | {"colsample_bytree": 0.7670777736372738, "learning_rate": 0.19276041730313465, "max_depth": 3, "n_estimators": 164, "subsample": 0.7816961974601263, "max_df": 0.6913515729152793, "max_features": 3109, "min_df": 4, "ngram_range": [1, 2]} | 0.294 | 0.328 |
| K-Nearest-Neighbor | Glove | {"vectorizer": glove-840B-300d, "leaf_size": 24, "n_neighbors": 5, "weights": "distance"} | 0.426 | 0.406 | {"vectorizer": glove-42B-300d, "leaf_size": 33, "n_neighbors": 4, "weights": "distance"} | 0.282 | 0.337 |
| Logistic Regression | | {"vectorizer": glove-42B-300d, "C": 6.999135822979566, "penalty": "l2", "solver": "lbfgs"} | 0.572 | 0.513 | {"vectorizer": glove-42B-300d, "C": 9.76042118194638, "penalty": "l2", "solver": "lbfgs"} | 0.340 | 0.335 |
| Random Forest | | {"vectorizer": glove-42B-300d, "bootstrap": false, "criterion": "gini", "max_depth": 95, "max_features": 0.2077430655332096, "min_samples_leaf": 1, "min_samples_split": 8, "n_estimators": 198} | 0.423 | 0.424 | {"vectorizer": glove-42B-300d, "bootstrap": false, "criterion": "entropy", "max_depth": 80, "max_features": 0.9425728949509299, "min_samples_leaf": 8, "min_samples_split": 17, "n_estimators": 326} | 0.263 | 0.231 |
| SVM | | {"vectorizer": glove-42B-300d, "C": 5.121773443923852, "degree": 4, "gamma": "auto", "kernel": "linear"} | 0.556 | 0.500 | {"vectorizer": glove-840B-300d, "C": 9.291710982638396, "degree": 5, "gamma": "auto", "kernel": "linear"} | 0.352 | 0.358 |
| XGBoost | | {"vectorizer": glove-840B-300d, "colsample_bytree": 0.557957280470468, "learning_rate": 0.23112312784449587, "max_depth": 4, "n_estimators": 455, "subsample": 0.6505946735495828} | 0.537 | 0.503 | {"vectorizer": glove-42B-300d, "colsample_bytree": 0.594405164158504, "learning_rate": 0.2845324775722167, "max_depth": 6, "n_estimators": 115, "subsample": 0.6025027592743828} | 0.300 | 0.283 |
| N | | | 667 | 1,000 | | 711 | 1,047 |

*Note.* ML-based-classifiers were trained with a dataset n=2,000 with a representative label distribution. Observation count (N) of macro averaged F1-score of cross-validated samples can vary depending on the data splits. On average, they are one-third of the training data size.

**Table A2 Hyperparameter configurations and performance evaluation for convolutional neural network for demonstration studies 1 & 2**

| | (1) Best Performing Hyperparameter Configuration | Best CV Macro F1 | HO Macro F1 | (2) Best Performing Hyperparameter Configuration | Best CV Macro F1 | HO Macro F1 |
|---|---|---|---|---|---|---|
| Convolutional Neural Network | {"batch_size": 60, "max_epochs": 12, "dense_units": 72, "dropout": 0.46777990469323594, "kernel_size": 6, "n_filters": 84, "trainable": true, "num_words": 16763, "seq_len": 214} | 0.464 | 0.484 | {"batch_size": 58, "max_epochs": 10, "dense_units": 72, "dropout": 0.40989527282572635, "kernel_size": 5, "n_filters": 228, "trainable": true, "num_words": 26815, "seq_len": 195} | 0.280 | 0.353 |
| N | | 667 | 1,000 | | 711 | 1,047 |

*Note.* CNN was trained with a dataset n=2,000 with a representative label distribution. Observation count (N) of macro averaged F1-score of cross-validated samples can vary depending on the data splits. On average, they are one-third of the training data size.

**Table A3 Hyperparameter configurations and performance evaluation for pre-trained language models for demonstration studies 1 & 2**

| | (1) Best Performing Hyperparameter Configuration | HO Macro F1 | (2) Best Performing Hyperparameter Configuration | HO Macro F1 |
|---|---|---|---|---|
| BERT | {"learning_rate": 2.3139649033257978e-05, "batch_size": 8, "num_train_epochs": 3, "weight_decay": 0.0072173948625224625} | 0.530 | {"learning_rate": 1.3205102630174912e-05, "batch_size": 8, "num_train_epochs": 5, "weight_decay": 0.0069800190183135335} | 0.288 |
| ELECTRA | {"learning_rate": 1.2868586517068187e-05, "batch_size": 8, "num_train_epochs": 5, "weight_decay": 0.004205974315203549} | 0.528 | {"learning_rate": 2.3139649033257978e-05, "batch_size": 8, "num_train_epochs": 3, "weight_decay": 0.0072173948625224625} | 0.254 |
| RoBERTa | {"learning_rate": 1.2868586517068187e-05, "batch_size": 8, "num_train_epochs": 5, "weight_decay": 0.004205974315203549} | 0.579 | {"learning_rate": 1.2868586517068187e-05, "batch_size": 8, "num_train_epochs": 5, "weight_decay": 0.004205974315203549} | 0.402 |
| XLNet | {"learning_rate": 1.2868586517068187e-05, "batch_size": 8, "num_train_epochs": 5, "weight_decay": 0.004205974315203549} | 0.547 | {"learning_rate": 1.3205102630174912e-05, "batch_size": 8, "num_train_epochs": 5, "weight_decay": 0.0069800190183135335} | 0.436 |
| N | | 1,000 | | 1,047 |

*Note.* PLMs were fine-tuned with a dataset n=2,000 with a representative label distribution.

**Figure C1 Additional Analyses on the Role of Training Data Distribution and Data Size**

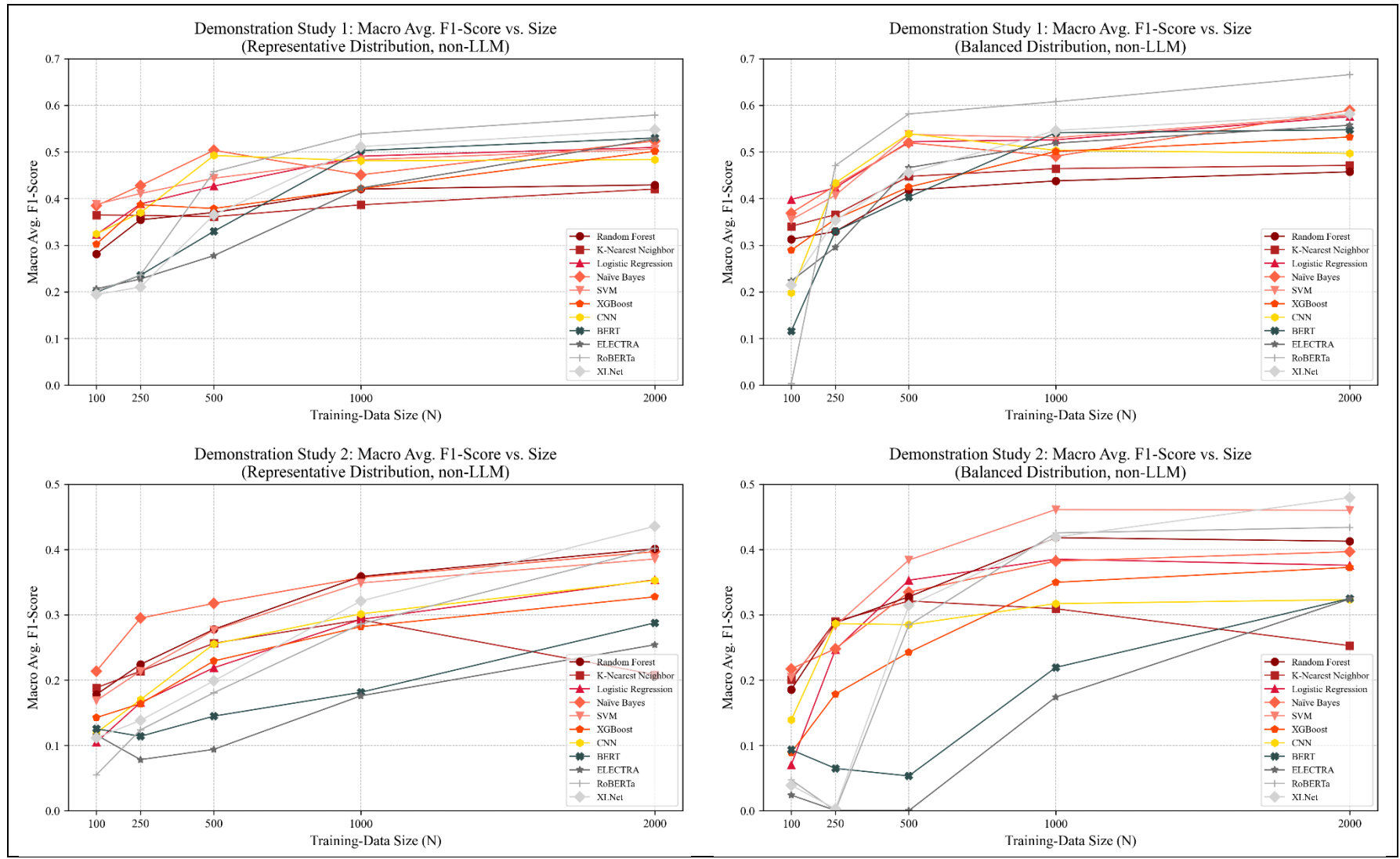

*Notes.* Embeddings for the ML classifiers were created using TF-IDF, and embeddings for the CNN were generated using Glove 42B-300d. Y-axes are scaled to fit best and worst model values.